\def\MDLMPECombined{1}
\documentclass[letterpaper]{article}
\usepackage[preprint]{mdlmpe2027}
\usepackage[hyphens]{url}
\usepackage{graphicx}
\def\UrlFont{\rm}
\usepackage{natbib}
\usepackage{caption}
\usepackage{booktabs}
\usepackage{multirow}
\usepackage{amsmath,amssymb,amsfonts}
\usepackage{bm}

\title{MDLMPE: Distribution Aware Positional Encoding for Masked Diffusion Language Models}
\author{
Tong Ling\textsuperscript{\rm 1},
Hang Lei\textsuperscript{\rm 2},
Feng Xiao\textsuperscript{\rm 2},
Changhui Sun\textsuperscript{\rm 3},
Jiahang Xie\textsuperscript{\rm 4},
Hao Liu\textsuperscript{\rm 2},
Lu Liu\textsuperscript{\rm 2},
Yanlong Du\textsuperscript{\rm 2}
}
\affiliations{
\textsuperscript{\rm 1}University Chinese Academic of Science\\
\textsuperscript{\rm 2}HUJING Digital Media \& Entertainment Group\\
\textsuperscript{\rm 3}State Key Laboratory for Novel Software Technology, Nanjing University, China\\
\textsuperscript{\rm 4}School of Data Science, Fudan University, Shanghai, China
}

\begin{document}

\maketitle

\begin{abstract}
Masked diffusion language models (MDLMs) enable parallel generation and bidirectional context modeling, but their positional context differs fundamentally from that of autoregressive (AR) models. Whereas AR decoding exposes a contiguous prefix, MDLM denoising produces dynamic, non-contiguous configurations of revealed and masked tokens. Conventional positional encodings such as RoPE capture sequence order and pairwise displacement but remain insensitive to this evolving token-availability structure. To address this limitation, we propose MDLMPE, a positional encoding designed specifically for masked diffusion. To the best of our knowledge, MDLMPE is the first method to make positional representations explicitly aware of the changing revealed/masked configuration. It represents token availability as a binary sequence, applies distance-aware Gaussian weighting, and projects the resulting pattern through a cosine basis to obtain distribution-aware positional features. These features are added to token embeddings and mapped by a lightweight MLP to angular offsets that modulate the standard RoPE phases. Extensive experiments on LLaDA and DREAM demonstrate that MDLMPE generally outperforms conventional positional encoding methods across supervised fine-tuning, pretraining, zero-shot evaluation, and block-diffusion settings. Further ablations show that the complete combination of availability state, Gaussian locality, spectral basis, and embedding injection yields the strongest result. These results establish the evolving token-availability distribution as a useful positional signal for masked diffusion language models.
\end{abstract}

\section{Introduction}

Autoregressive (AR) large language models~\cite{radford2019language,brown2020language,touvron2023llama} have transformed language modeling, yet their left-to-right factorization imposes sequential generation and causal, unidirectional attention. Bidirectional pretraining enriches contextual representations, but AR decoding remains sequential. This limitation motivates masked diffusion language models (MDLMs), which iteratively denoise a masked sequence and support parallel generation, flexible generation orders~\cite{stern2019insertion}, and bidirectional context modeling~\cite{nie2025llada,ye2025dream,seeddiffusion2025}. Systems including LLaDA~\cite{nie2025llada}, DREAM~\cite{ye2025dream}, and LLaDA2.0~\cite{bie2025llada2} show that this paradigm can scale to billions of parameters while retaining competitive language-modeling ability. As scalability becomes established, a more specific question emerges: how should positional context be represented when the visible sequence state changes throughout denoising?

\begin{figure*}[t]
\centering
\includegraphics[width=0.85\textwidth]{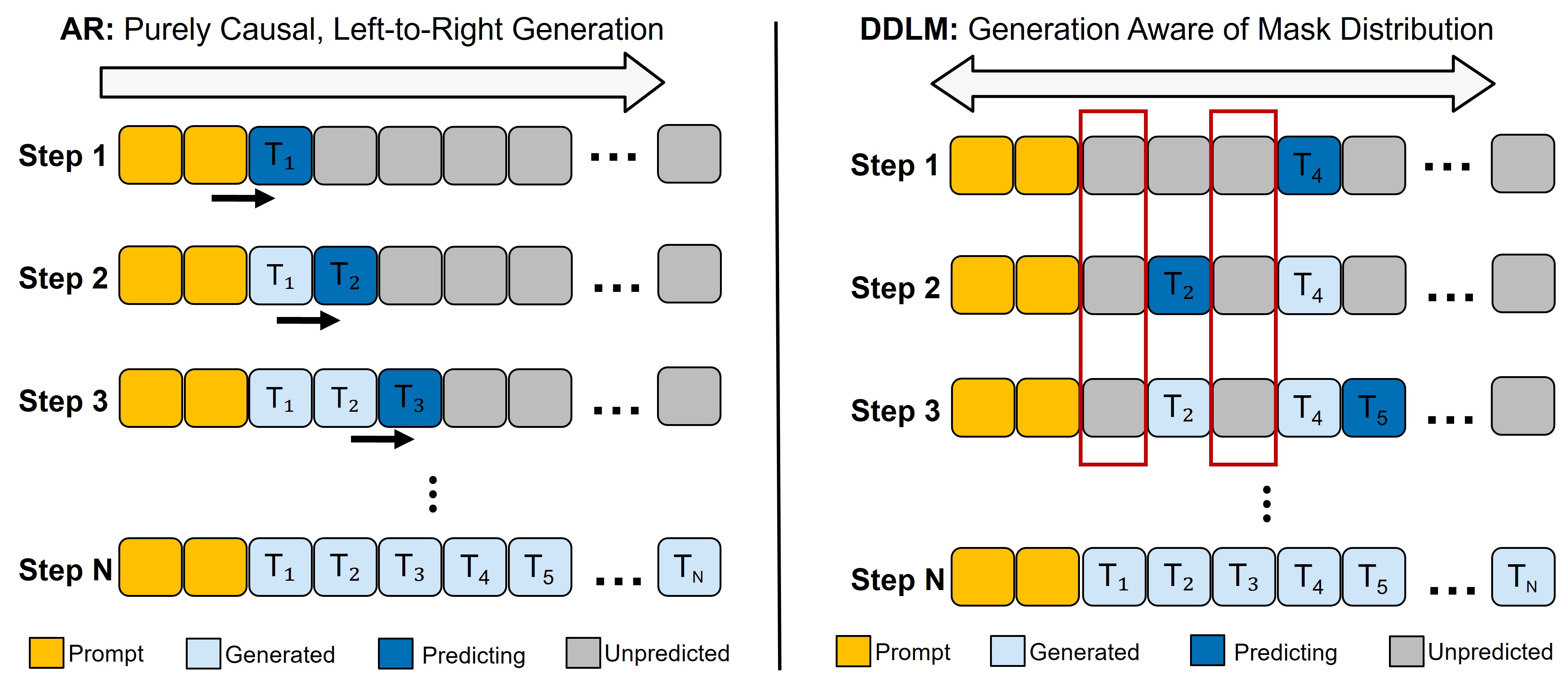}
\caption{Position availability in AR and masked diffusion generation. AR exposes a contiguous prefix, whereas MDLM denoising reveals a changing, non-contiguous set of response positions.}
\label{fig:ar_MDLM}
\end{figure*}

This question arises because MDLM generation exposes a positional structure that AR decoding never encounters. An AR model reveals a contiguous prefix, whereas an MDLM may recover tokens in a non-contiguous order. A token can retain the same absolute index while the semantic evidence available around it changes as other positions are revealed or re-masked. Its prediction depends not only on where it is located, but also on how the available context is organized. Figure~\ref{fig:ar_MDLM} shows this contrast by comparing the monotonic AR prefix with the evolving position availability of MDLM generation. As highlighted by the two red boxes, the two masked tokens remain at fixed sequence positions throughout denoising, while the distribution of surrounding tokens continually changes; existing positional encoding methods cannot reflect this evolving contextual structure.

\begin{figure*}[t]
\centering
\includegraphics[width=0.875\textwidth]{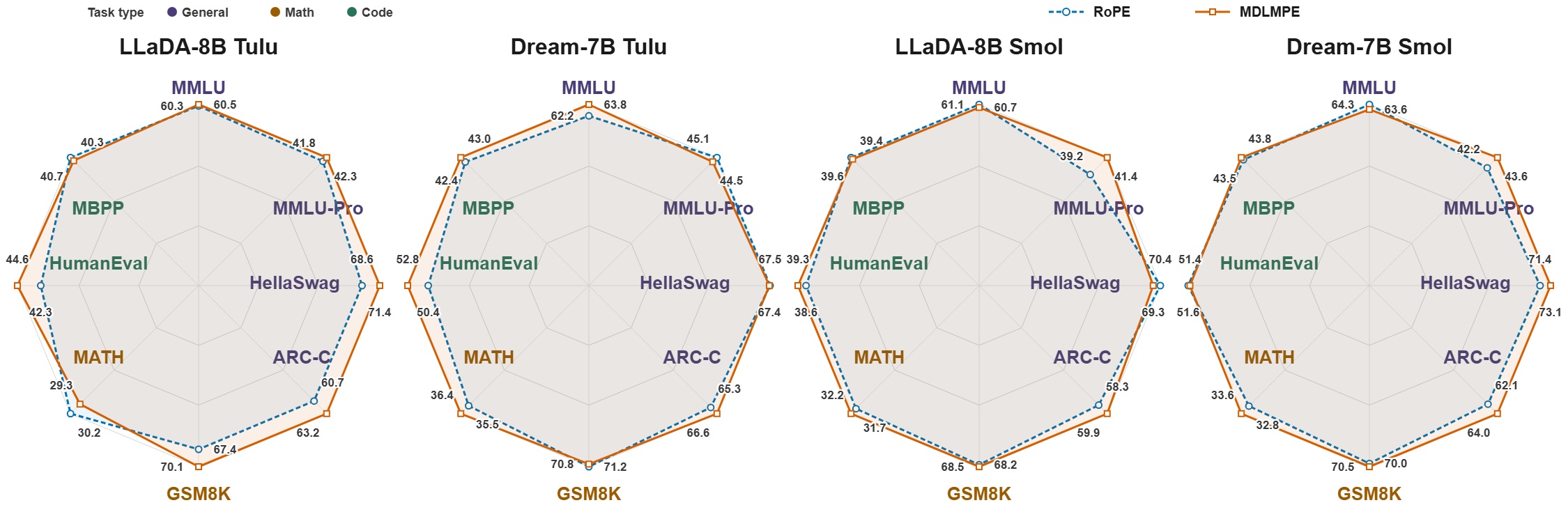}
\caption{Instruction post-training accuracies for the four table settings, ordered left to right as LLaDA-8B Tulu, DREAM-7B Tulu, LLaDA-8B SmolTalk, and DREAM-7B SmolTalk. Dashed blue circles denote RoPE and solid orange squares denote MDLMPE.}
\label{fig:large_sft_overview}
\end{figure*}

The limitation becomes clearer when viewed through Rotary Position Embedding (RoPE)~\cite{su2024roformer}, the mainstream positional encoding used by MDLM backbones. RoPE associates each index with a rotation and represents the relation between two positions through their distance-dependent phase difference. This geometry is suited to AR generation: every visible prefix position contains a real token, so pairwise displacement provides a reliable description of where usable context is located. In an MDLM, however, the same indices and displacement may occur under different revealed/masked configurations. The attention relation should therefore reflect both pairwise distance and the distribution of available evidence around the two positions. This observation suggests three requirements for a position-distribution representation: (i) revealed-token density, which estimates the amount of semantic evidence; (ii) target-relative distance to the revealed-token distribution, which indicates contextual relevance; and (iii) the global topology of the revealed/masked pattern, which distinguishes distributional structures that isolated indices cannot express.

This objective also distinguishes MDLMPE from RoPE adaptations for long-context diffusion models~\cite{liu2025longllada,he2025ultrallada}. LongLLaDA applies NTK-based extrapolation, while UltraLLaDA modifies RoPE during post-training for 128K contexts. These methods address coordinate behavior beyond the pretraining length; our focus is the revealed/masked availability pattern within the current denoising sequence.

Motivated by this gap, we propose \textbf{MDLMPE}, a positional encoding that makes the denoising state an explicit positional signal. MDLMPE represents real and masked tokens as a binary availability sequence. It then applies target-centered Gaussian weighting so that the representation reflects where revealed evidence lies relative to each token; for an attention pair, the two centered views are averaged into a joint distribution. A cosine-basis projection converts this pattern into distribution-aware features. MDLMPE uses them through two complementary paths: token features are added to embeddings, while a lightweight MLP maps pair features to angular offsets for RoPE. The resulting design preserves RoPE's relative-position geometry while making both token representations and attention rotations responsive to the evolving mask pattern.

To determine whether this positional signal is useful both when adapting existing checkpoints and when learning from scratch, we evaluate MDLMPE on LLaDA and DREAM. The matched 7B/8B post-training results summarized in Figure~\ref{fig:large_sft_overview} show improvements across most knowledge, reasoning, and code tasks on both Tulu and SmolTalk, with broadly comparable performance on the remaining tasks. This consistency across model architectures and training datasets motivates our subsequent controlled evaluations, which cover 100M-parameter pretraining and zero-shot evaluation, block diffusion, and component ablations. Our contributions are threefold:

\begin{itemize}
    \item We introduce MDLMPE, a positional encoding for MDLMs that explicitly models the evolving position-distribution structure induced by the revealed/masked token configuration.
    \item We identify the changing token-availability pattern as a distinctive challenge of MDLM decoding and construct a Gaussian-weighted, cosine-projected representation of this pattern.
    \item Extensive experiments on LLaDA and DREAM demonstrate improvements over conventional positional encodings, while controlled ablations characterize how the MDLMPE components combine.
\end{itemize}

\section{Related Work}

\paragraph{Positional encoding.}
Position encoding provides the coordinate system through which attention distinguishes order~\cite{vaswani2017attention}. Existing designs instantiate this role in different ways: absolute encodings attach vectors to indices~\cite{devlin2019bert}; relative methods modify attention by pairwise displacement~\cite{shaw2018self,dai2019transformerxl,he2021deberta}; RoPE rotates queries and keys~\cite{su2024roformer}; and ALiBi adds a head-specific distance bias~\cite{press2022alibi}. Position Interpolation and YaRN further extend pretrained coordinates to longer contexts~\cite{chen2023extending,peng2024yarn}.

Although these methods differ in form, they share a fixed-coordinate premise: an index is mapped to a vector, rotation, or bias independently of the current observation state. Extrapolation methods modify the rotary spectrum or training positions---xPos changes the parameterization~\cite{sun2023xpos}, LongRoPE uses non-uniform rescaling~\cite{ding2024longrope}, and PoSE trains on skipped indices~\cite{zhu2024pose}---but retain this premise. LongLLaDA and UltraLLaDA likewise adapt RoPE for long-context diffusion models~\cite{liu2025longllada,he2025ultrallada}; MDLMPE addresses a different question by conditioning an in-window position on the observed availability state.

This distinction is largely hidden in AR decoding, where the causal mask and revealed prefix coincide, but becomes consequential in an MDLM that exposes both tokens and masks. An index-only encoding states that two positions are $k$ steps apart without indicating whether either has nearby usable evidence. MDLMPE therefore retains RoPE's index relation while adding the observed/masked configuration to the residual stream and rotary coordinates.

\paragraph{Masked diffusion language modeling.}
In parallel, diffusion language modeling has progressed from extending continuous denoising~\cite{ho2020denoising,nichol2021improved,song2021scorebased} to categorical states~\cite{austin2021d3pm,hoogeboom2021argmax} toward increasingly capable text generators. Diffusion-LM, DiffuSeq, and iterative masked generation establish text settings~\cite{li2022diffusionlm,gong2023diffuseq,ghazvininejad2019maskpredict,gu2019levenshtein,chang2022maskgit}; SEDD, MD4, RADD, and TESS refine objectives or parameterizations~\cite{lou2024sedd,shi2024md4,ou2025radd,mahabadi2024tess}; and LLaDA, DREAM, LLaDA2.0, and Seed Diffusion scale the approach~\cite{nie2025llada,ye2025dream,bie2025llada2,seeddiffusion2025}.

Together, these advances improve diffusion objectives, scale, and generation strategies, yet the denoiser still repeatedly infers missing tokens from a partially observed sequence. Diffusion time is usually encoded globally, whereas the spatial layout of the observed subset remains implicit in mask-token activations and attention. DiffusionBERT, Block Diffusion, and DFlash modify the objective, generation factorization, or speculative-drafting mechanism~\cite{he2023diffusionbert,arriola2025blockdiffusion,chen2026dflash}. MDLMPE complements these directions: it leaves their generation machinery unchanged and adds a coordinate derived from the realized availability pattern.

\section{Method}
\label{sec:method}

The preceding discussion identifies a representation gap: masked denoising needs positional signals that follow the changing mask state, distinguish both the amount and location of available evidence, and preserve the geometry of a RoPE-pretrained backbone. MDLMPE meets these requirements with a shared availability encoder and two complementary paths: token-centered features enrich the residual stream, while pair-centered features adjust selected rotary relations.

\subsection{Availability-Aware Positional Representation}

The first requirement is state sensitivity. Although an MDLM can identify individual \texttt{[MASK]} tokens, their spatial arrangement remains entangled with content. We expose this structure explicitly through a minimal content-free state. For $\bm{x}=(x_1,\ldots,x_L)$, define
\begin{equation}
b_t=\mathbb{I}[x_t\neq x_{\mathrm{mask}}]a_t, \qquad b_t\in\{0,1\},
\label{eq:main_binary}
\end{equation}
where $a_t\in\{0,1\}$ excludes attention-invalid positions. The signal therefore changes as soon as a token is revealed, without encoding lexical content or clean targets.

State alone, however, does not express how strongly each revealed position should influence a target. Because nearby and distant evidence need not be equally useful, a normalized truncated Gaussian converts the global mask pattern into a target-relative neighborhood:
\begin{equation}
w_{i,t}=\frac{\exp\left(-\frac{(i-t)^2}{2\sigma^2}\right)\mathbb{I}[|i-t|\leq W]a_t}{\sum_{u=1}^{L}\exp\left(-\frac{(i-u)^2}{2\sigma^2}\right)\mathbb{I}[|i-u|\leq W]a_u+\varepsilon},
\label{eq:main_gaussian}
\end{equation}
where $W$ is the radius and $\sigma=\max(\rho W,1)$. Normalization stabilizes boundaries; truncation controls locality and cost. We use $W/L=3/16$ and $\rho=1/4$ (e.g., $W=192$, $\sigma=48$ at $L=1024$). The symmetric kernel models bidirectional evidence without a learned left/right preference.

\begin{figure}[!ht]
\centering
\includegraphics[width=0.98\columnwidth]{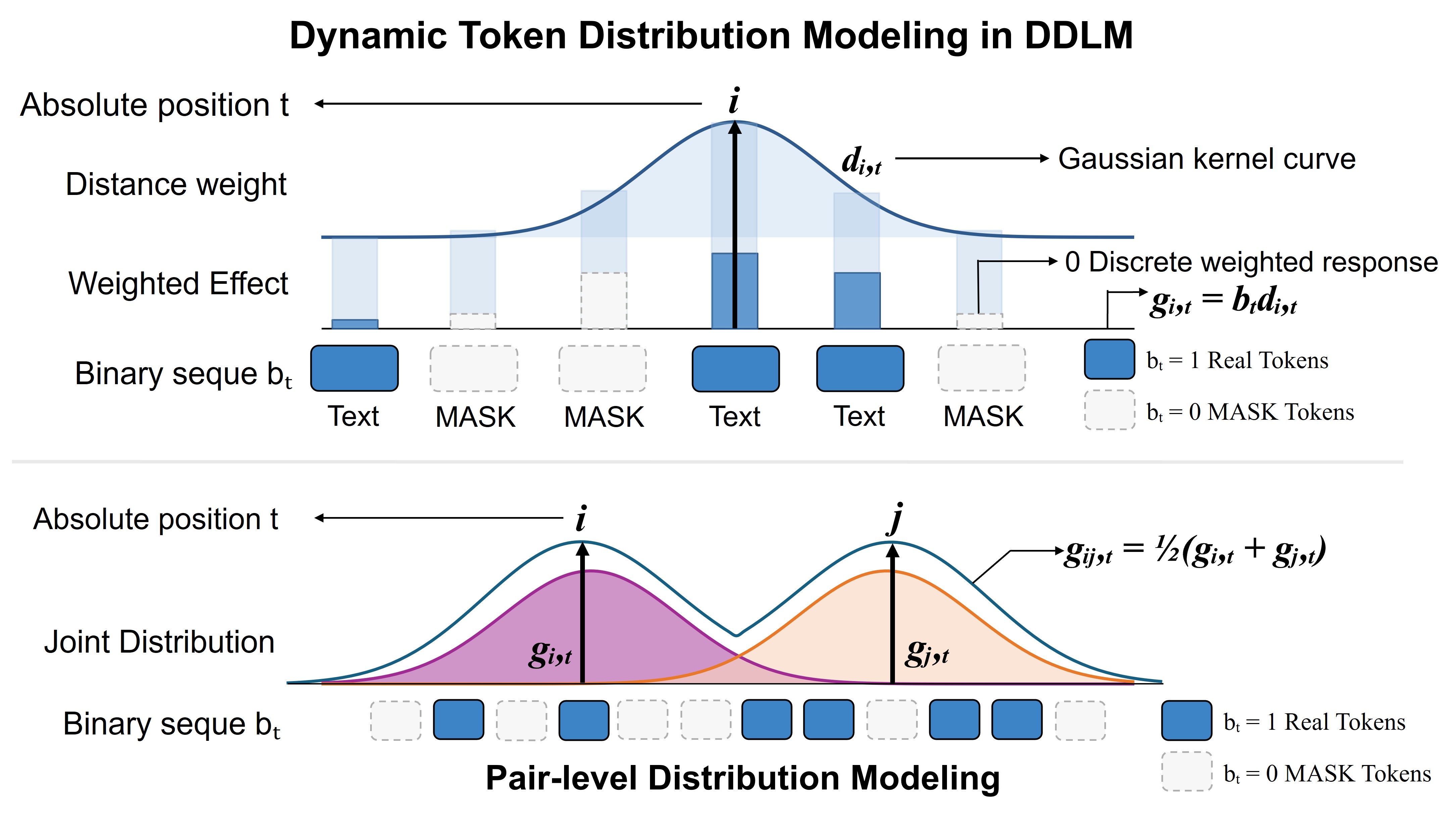}
\caption{Gaussian local encoding for one position (top) and the pair-level distribution used by the rotary branch (bottom). The token-centered upper construction supplies the embedding branch. For an attention pair $(i,j)$, the implemented rotary branch averages the two weighted sequences as $g_{ij,t}=\tfrac12(g_{i,t}+g_{j,t})$.}
\label{fig:local_pair}
\end{figure}

\begin{figure}[!ht]
\centering
\includegraphics[width=0.95\columnwidth]{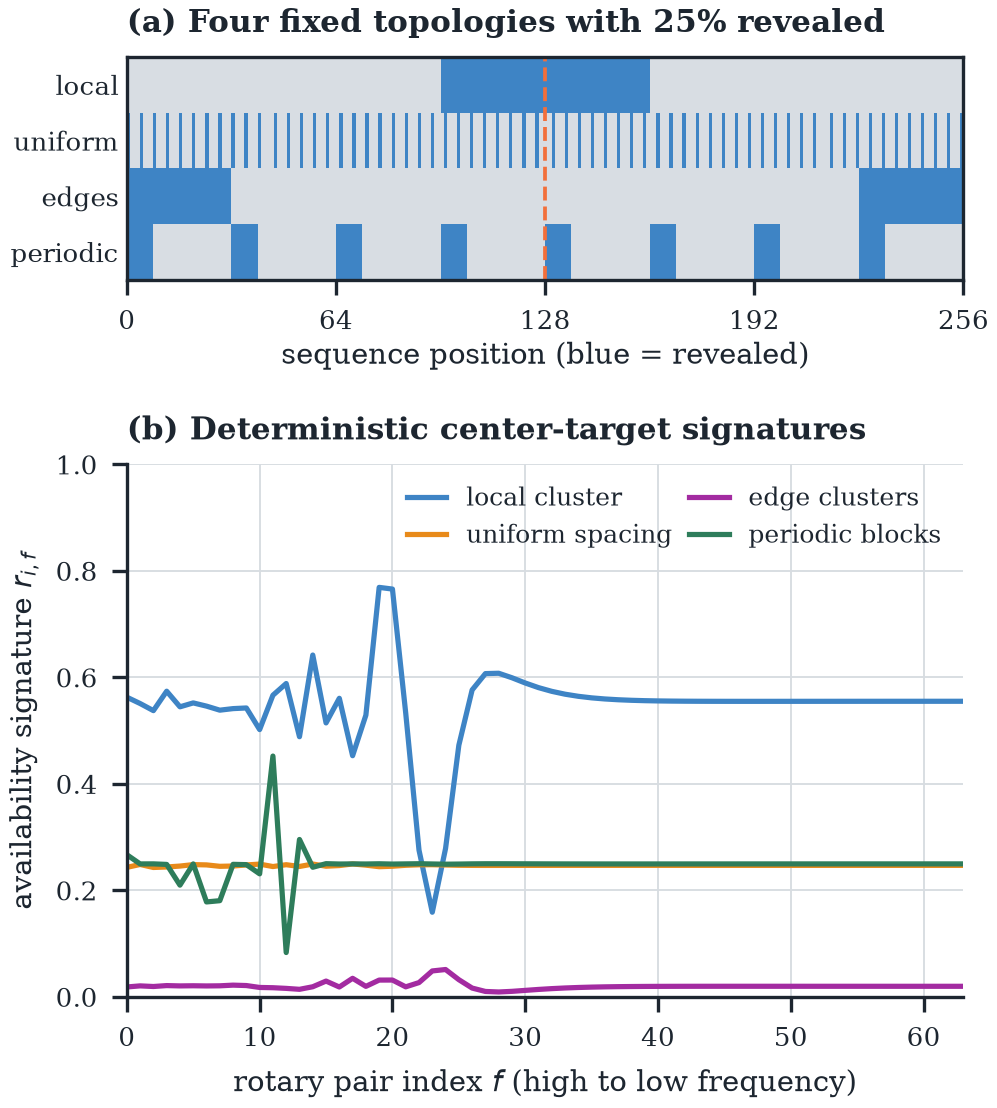}
\caption{Equal-density availability patterns produce different deterministic MDLMPE signatures under the default geometry. Each length-256 mask reveals exactly 64 positions (25\%); the dashed line marks the center target. With $W/L=3/16$, $\sigma/W=1/4$, and a fixed RoPE frequency grid, local, uniformly spaced, edge-clustered, and periodic arrangements yield distinct vectors $\{r_{i,f}\}_f$. The figure illustrates representational topology, not learned embeddings or benchmark outcomes.}
\label{fig:canonical_topologies}
\end{figure}

Local weighting makes the representation target-relative, but a weighted count would still collapse equal-density layouts. We therefore preserve their topology on RoPE's multiscale frequency grid. For $F=d_h/2$ rotary pairs, let $c_{t,f}=\frac12[1+\sin(t\omega_f)]$ be a nonnegative shifted-sine coordinate and form
\begin{equation}
\begin{aligned}
u_{i,f}&=\sum_{t=1}^{L}w_{i,t}b_tc_{t,f},\\
d_{i,f}&=\sum_{t=1}^{L}w_{i,t}a_tc_{t,f}+\varepsilon,\qquad
r_{i,f}=\frac{u_{i,f}}{d_{i,f}}.
\end{aligned}
\label{eq:main_projection}
\end{equation}
with $0\le r_{i,f}\le1$. Low frequencies summarize broad availability and higher frequencies retain finer changes, distinguishing the equal-density layouts in Figure~\ref{fig:canonical_topologies}.

Once this topology has been encoded, it must enter the model at the granularity required by each path. Let $\bm u_i=(u_{i,1},\ldots,u_{i,F})$ and $\bm d_i=(d_{i,1},\ldots,d_{i,F})$ denote the token-centered availability numerator and its all-valid reference. Token embeddings use $\bm z_i=\bm W_p\bm u_i+\bm b_p$ and $\widetilde{\bm e}_i=\bm e_i+\gamma\bm z_i$, with $\gamma\le0.10$ initialized to $0.01$. For the pair branch, define $\bm A_i\equiv\bm u_i$ and $\bm D_i\equiv\bm d_i$. The symmetric pair feature is
\begin{equation}
\bm r_{ij}=g(\bm A_i,\bm D_i,\bm A_j,\bm D_j),
\qquad
[\bm r_{ij}]_f=\frac{A_{i,f}+A_{j,f}}{D_{i,f}+D_{j,f}},
\label{eq:main_pair_feature}
\end{equation}
which is equivalent to $u_{ij,f}=\tfrac12(u_{i,f}+u_{j,f})$, $d_{ij,f}=\tfrac12(d_{i,f}+d_{j,f})$, and $r_{ij,f}=u_{ij,f}/d_{ij,f}$ in Figure~\ref{fig:local_pair}. Thus $\bm u_i$ describes one token's surroundings, whereas $\bm r_{ij}$ summarizes the joint context used by an attention pair. Appendices~A and F provide safeguards and deterministic analyses.

\subsection{Availability-Conditioned Rotary Adaptation}

\begin{figure*}[t]
\centering
\includegraphics[width=0.8\textwidth]{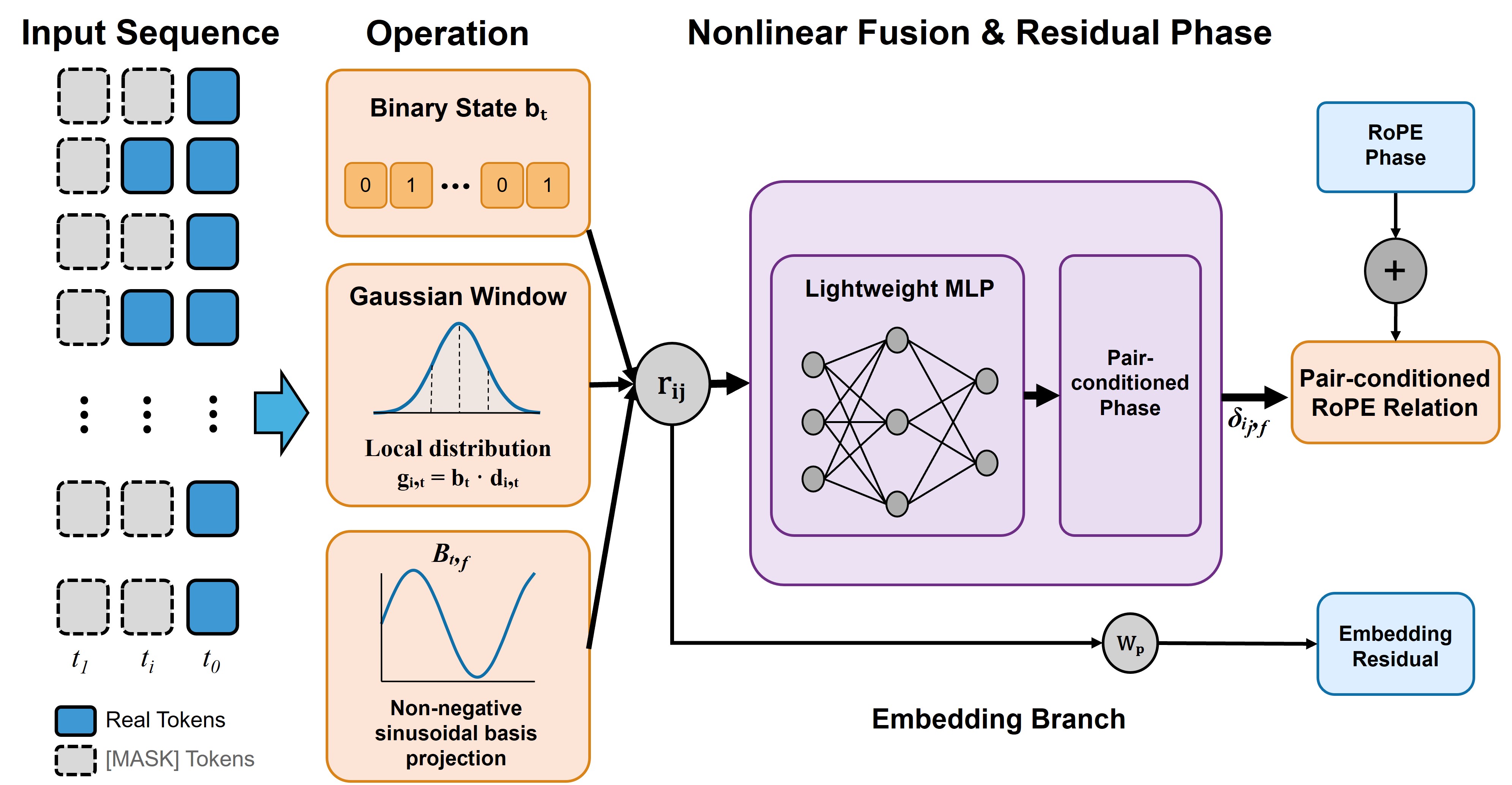}
\caption{System-level view of the high-level two-branch architecture. The token feature $\bm u_i$ supplies the enabled embedding branch, while the pair feature $\bm r_{ij}$ supplies the shared MLP and bounded RoPE residual; half of the interleaved rotary pairs remain ordinary RoPE. The pair feature $\bm r_{ij}$, its MLP transformation $\bm\psi_{ij}$, and the resulting phase residual $\delta_{ij,f}$ are defined in the main text.}
\label{fig:pipeline}
\end{figure*}

The pair feature now captures joint context, but replacing RoPE angles with it would discard pretrained displacement geometry. MDLMPE instead uses this information as a bounded residual, beginning with the fixed angularization
\begin{equation}
\bm s_{ij}=\arccos(2\bm r_{ij}-\bm1)-\tfrac{\pi}{2}\bm1,
\label{eq:main_injection}
\end{equation}
where $\arccos$ is applied elementwise. The inverse-cosine map spreads the normalized availability range over a stable angular domain, while the subsequent learnable transformation determines how this signal should interact across frequencies; $\bm s_{ij}$ is therefore an intermediate coordinate rather than a rotary angle. To make the high-level path in Figure~\ref{fig:pipeline} explicit, a shared hidden layer applies a linear transformation and SiLU activation to $\bm s_{ij}$, producing $\bm\psi_{ij}\in\mathbb R^H$, where $H=\max(16,F/2)$ by default. Each frequency-specific output head then maps $\bm\psi_{ij}$ to the phase residual $\delta_{ij,f}=h_f(\bm\psi_{ij})$:
\begin{equation}
h_f(\bm\psi_{ij})
=\tau_{\max}\tanh(\bm w_{2,f}^{\top}\bm\psi_{ij}+b_{2,f}).
\label{eq:main_phase_residual}
\end{equation}
Here $\tau_{\max}=0.25$, $\bm w_{2,f}^{\top}$ is the $f$th row of the shared output layer, and the output heads are zero initialized. This gives the high-level computation $\bm r_{ij}\!\rightarrow\!\bm\psi_{ij}\!\rightarrow\!\delta_{ij,f}$. Let $\bar{\bm q}_{i,f}=\bm R(i\omega_f)\bm q_{i,f}$ and $\bar{\bm k}_{j,f}=\bm R(j\omega_f)\bm k_{j,f}$ denote the ordinary RoPE-rotated two-dimensional query and key pair. The adapted pair contribution is
\begin{equation}
S_{ij,f}=\begin{cases}
\bar{\bm q}_{i,f}^{\top}\bar{\bm k}_{j,f}, & f\in\mathcal F_{\mathrm{RoPE}},\\
\bar{\bm q}_{i,f}^{\top}\bm R(\delta_{ij,f})\bar{\bm k}_{j,f}, & f\in\mathcal F_{\mathrm{MDLMPE}},
\end{cases}
\label{eq:main_final_angle}
\end{equation}
with $|\mathcal F_{\mathrm{MDLMPE}}|\approx F/2$ and $S_{ij}=\sum_f S_{ij,f}/\sqrt{d_h}$. Equivalently, the angular relation used by attention is
\begin{equation}
\Delta\phi_{ij,f}=\begin{cases}
(j-i)\omega_f, & f\in\mathcal{F}_{\mathrm{RoPE}},\\
(j-i)\omega_f+\delta_{ij,f}, & f\in\mathcal{F}_{\mathrm{MDLMPE}}.
\end{cases}
\label{eq:main_pair}
\end{equation}
The construction thus preserves the familiar displacement relation and augments it only where the observed context provides additional structure. Every adapted pair retains RoPE's displacement term while adding joint-context information; clipping $\arccos$ prevents endpoint singularities, and bounding $\bm\delta_{ij}$ limits the perturbation.

\subsection{Checkpoint-Compatible Integration}

The preceding construction is useful in large backbones only if its additional expressivity can be introduced without disturbing the pretrained interface. LLaDA and DREAM therefore share one module that builds availability from token IDs and the validity mask, computes Gaussian summaries once per forward pass, and reuses them across layers. Hidden size, tokenizer, denoising schedule, and output head are unchanged.

Because pair-conditioned phases are evaluated inside attention, fused kernels require an MDLMPE-aware extension; the reference implementation evaluates these phases in query blocks. Local aggregation costs $O(LWF)$, the enabled embedding projection costs $O(LFd)$, and pair-phase evaluation adds $O(L^2FH)$ for the shared map before the selected-frequency score update. A query block of size $C$ bounds extra working storage by $O(CLF)$.

\paragraph{Frequency and head allocation.}
Within this implementation, one availability representation is shared across the model rather than divided into four view groups. Rotary pairs alternate between RoPE and MDLMPE, giving an approximately 50/50 split: unchanged pairs preserve a direct checkpoint path, while interleaving covers the frequency range. Training from scratch and inference activate all key--value heads. During instruction post-training, the active fraction rises linearly from zero at 10\% of updates to one at 90\%; selected heads are distributed approximately uniformly, and grouped query heads follow their key--value head.

\paragraph{Implementation details.}
Numerical safeguards preserve this gradual integration in practice. Availability is constructed before rotary embedding and excludes invalid positions. The basis and Gaussian are fixed and symmetric; denominators and $\arccos$ inputs are clamped for mixed precision. Zero-initializing the phase-MLP output starts the rotary path exactly at RoPE, while the bounded embedding gate controls the second residual path.

Finally, the availability signal is always recomputed from the information actually visible to the denoiser. During pretraining, $\bm b$ is obtained from the sampled corruption state; during instruction post-training, it includes the prompt and revealed response tokens but excludes masked response targets. The same operation is repeated after each inference update. Training and inference therefore expose MDLMPE to the same information pattern, preventing the positional feature from leaking clean responses.

\begin{table*}[!t]
\centering
\small
\begin{tabular}{lcc|cc|cc|cc}
\toprule
\textbf{Metric} & \multicolumn{4}{c|}{\textbf{LLaDA-8B}} & \multicolumn{4}{c}{\textbf{DREAM-7B}} \\
\cmidrule(lr){2-5}\cmidrule(lr){6-9}
 & T RoPE & T MDLMPE & S RoPE & S MDLMPE & T RoPE & T MDLMPE & S RoPE & S MDLMPE \\
\midrule
MMLU (5)       & 60.3 & \textbf{60.5} & \textbf{61.1} & 60.7 & 62.2 & \textbf{63.8} & \textbf{64.3} & 63.6 \\
MMLU-Pro (0)   & 41.8 & \textbf{42.3} & 39.2 & \textbf{41.4} & \textbf{45.1} & 44.5 & 42.2 & \textbf{43.6} \\
HellaSwag (0)  & 68.6 & \textbf{71.4} & \textbf{70.4} & 69.3 & \textbf{67.5} & 67.4 & 71.4 & \textbf{73.1} \\
ARC-C (0)      & 60.7 & \textbf{63.2} & 58.3 & \textbf{59.9} & 65.3 & \textbf{66.6} & 62.1 & \textbf{64.0} \\
\midrule
GSM8K (4)      & 67.4 & \textbf{70.1} & 68.2 & \textbf{68.5} & \textbf{71.2} & 70.8 & 70.0 & \textbf{70.5} \\
MATH (4)       & \textbf{30.2} & 29.3 & 31.7 & \textbf{32.2} & 35.5 & \textbf{36.4} & 32.8 & \textbf{33.6} \\
\midrule
HumanEval (0)  & 42.3 & \textbf{44.6} & 38.6 & \textbf{39.3} & 50.4 & \textbf{52.8} & \textbf{51.6} & 51.4 \\
MBPP (4)       & \textbf{40.7} & 40.3 & \textbf{39.6} & 39.4 & 42.4 & \textbf{43.0} & 43.5 & \textbf{43.8} \\
\bottomrule
\end{tabular}
\caption{Instruction post-training results on Tulu (T) and SmolTalk (S), transposed for metric-wise comparison. Every RoPE/MDLMPE column pair starts from the same RoPE-pretrained checkpoint.}
\label{tab:large_sft}
\end{table*}

\subsection{Compatibility with Blockwise Denoising}
\label{sec:block_method}

Having defined the encoder independently of a particular sampler, we next consider whether the same representation remains meaningful under blockwise generation. If $q_s$ is the active-block boundary and $\mathcal P$ the prompt, blocked context exposes $\mathcal P\cup\{t:t\le q_s\}$, whereas the LLaDA-style regime exposes the full partially denoised response. MDLMPE encodes no block ID; it recomputes $\bm b$ from the state presented to attention and applies the same dual-path construction. It therefore supports both factorizations without changing the objective, schedule, or update rule.

\section{Experiments}
\label{sec:experiments}

\subsection{Experimental Design}

The method makes claims at several levels---checkpoint compatibility, from-scratch positional bias, robustness to changing visibility, and component interaction---so we organize the evaluation around these four questions. For checkpoint adaptation, we conduct instruction post-training of LLaDA-8B and DREAM-7B on Tulu-3 and SmolTalk, using supervised fine-tuning (SFT) as the optimization recipe~\cite{lambert2024tulu3,allal2025smollm2smolgoesbig}. To isolate the positional inductive bias, we pretrain matched 100M-parameter models on OpenWebText (OWT) for 200K steps at length 1{,}024~\cite{gokaslan2019openwebtext}. Component analyses use LLaDA-7B with 50K/2K Tulu train/test examples. All training uses $8\times$ A100 80GB GPUs.

To make these questions comparable, RoPE~\cite{su2024roformer} serves as the primary baseline, while ALiBi~\cite{press2022alibi} and absolute embeddings provide controlled-pretraining diagnostics. Within each comparison, tokenizer, corruption, token budget, optimizer, decoding, and normalization are fixed. Post-training activates MDLMPE heads progressively from 10\% to 90\% of updates; from-scratch models use it throughout. Appendix~B specifies the complete protocol.

\subsection{Instruction Post-Training at 7B/8B Scale}

With the evaluation protocol fixed, we first ask whether a state-dependent coordinate can be introduced after pretraining while preserving the backbone's learned geometry. The matched results in Table~\ref{tab:large_sft} show improvements across knowledge, reasoning, and code tasks. On LLaDA--Tulu, HellaSwag increases from 68.6 to 71.4 and GSM8K from 67.4 to 70.1. On DREAM--Tulu, MATH increases from 35.5 to 36.4 and HumanEval from 50.4 to 52.8; DREAM--SmolTalk also improves HellaSwag from 71.4 to 73.1. Since each comparison shares its starting checkpoint, data order, scheduler, decoding budget, and selection rule, these gains demonstrate that MDLMPE can add availability structure effectively through instruction post-training.

\subsection{Controlled Pretraining and Cross-Corpus Evaluation}

Post-training establishes checkpoint compatibility, but from-scratch pretraining asks the complementary question of whether availability awareness is useful as an inductive bias in its own right. We therefore train matched 100M-parameter models and report pretraining-validation and zero-shot perplexity in Table~\ref{tab:zero_shot}. For LLaDA-100M, MDLMPE obtains 229.80 on 1BW compared with 239.34 for RoPE, 203.65 on Text8 compared with 214.31, and 26.13 on OWT compared with 27.08. DREAM-100M shows particularly clear improvements on PTB, from 97.05 to 81.24, and on WikiText-2, from 63.74 to 58.82; its WikiText-103 perplexity also reaches 61.64. Together, the in-domain and external-corpus results show that availability-aware positioning contributes during pretraining and carries into zero-shot evaluation across different document structures.
\begin{table}[!ht]
\centering
\footnotesize
\begin{tabular}{lrrr}
\toprule
\textbf{Corpus} & \textbf{RoPE} & \textbf{ALiBi} & \textbf{MDLMPE} \\
\midrule
\multicolumn{4}{c}{\textbf{LLaDA-100M}} \\
\midrule
1BW      & 239.34 & 273.42 & \textbf{229.80} \\
PTB      & 144.55 & \textbf{135.60} & 170.18 \\
Text8    & 214.31 & 243.30 & \textbf{203.65} \\
WT103    & \textbf{103.12} & 107.09 & 104.84 \\
WT2      & 71.07  & 87.18  & \textbf{68.10} \\
\textbf{OWT}      & 27.08 & 26.41  & \textbf{26.13} \\
\midrule
\multicolumn{4}{c}{\textbf{DREAM-100M}} \\
\midrule
1BW      & 196.50 & 198.73 & \textbf{194.54} \\
PTB      & 97.05  & 101.53 & \textbf{81.24} \\
Text8    & 116.94 & \textbf{103.46} & 107.10 \\
WT103    & 64.49  & 62.79  & \textbf{61.64} \\
WT2      & 63.74  & 61.28  & \textbf{58.82} \\
\textbf{OWT}      & 52.37  & 52.64  & \textbf{52.29} \\
\bottomrule
\end{tabular}
\caption{Controlled pretraining-validation and zero-shot perplexity for 100M-parameter LLaDA and DREAM models. OWT is the training distribution; the remaining corpora evaluate cross-corpus transfer.}
\label{tab:zero_shot}
\end{table}

\subsection{Robustness under Blockwise Post-Training}
\label{sec:block_experiment}

\begin{figure*}[!t]
\begin{minipage}[t]{0.48\textwidth}
\vspace{0pt}
\centering
\includegraphics[width=0.96\linewidth]{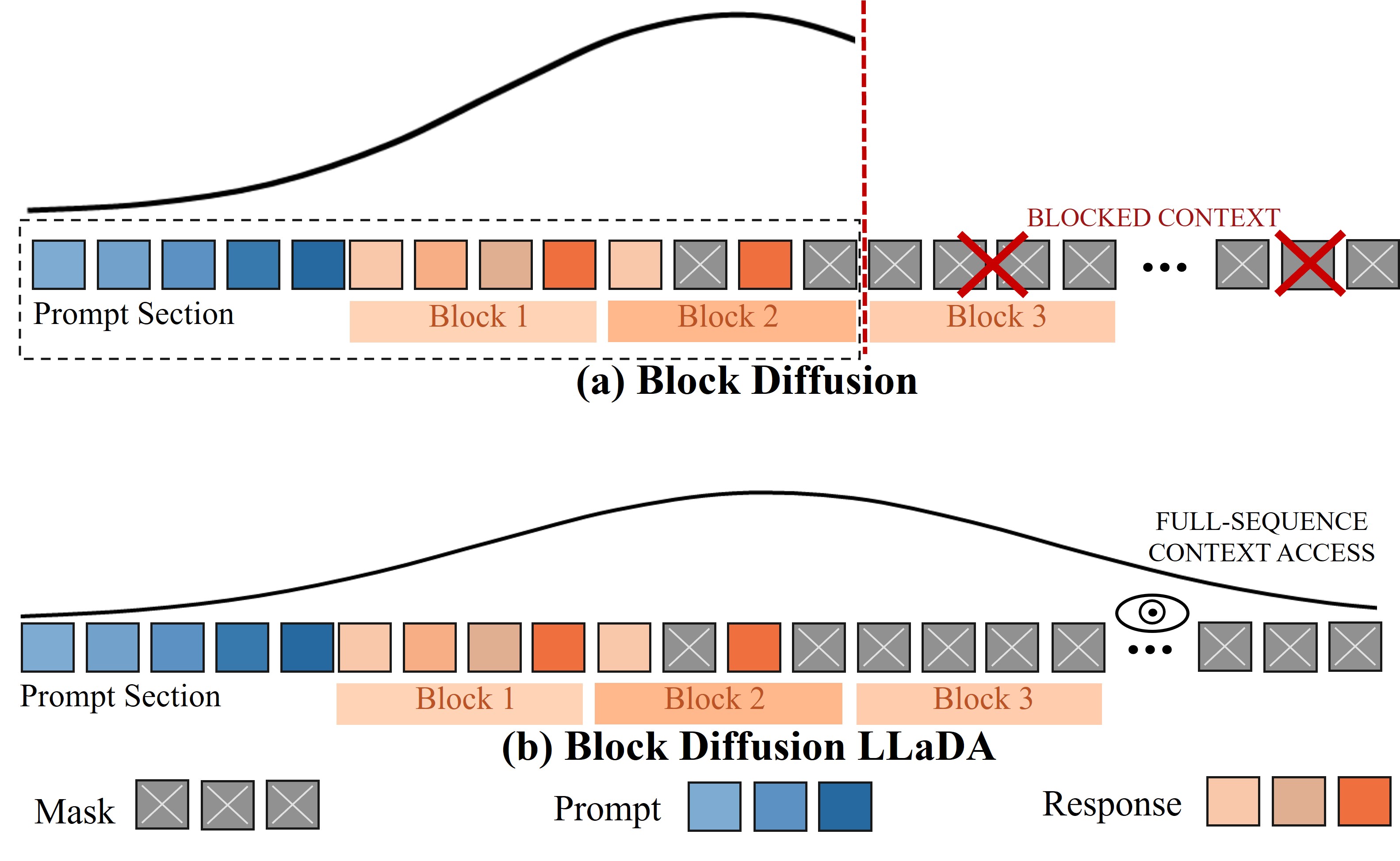}
\captionof{figure}{Blocked-context and full-sequence block diffusion. MDLMPE encodes the visibility pattern presented to attention.}
\label{fig:block_access}
\end{minipage}
\hfill
\begin{minipage}[t]{0.48\textwidth}
\vspace{0pt}
\centering
\small
\begin{tabular}{lcrrr}
\toprule
\textbf{Pos.} & \textbf{B} & \textbf{MMLU} & \textbf{ARC} & \textbf{Hella.} \\
\midrule
\multicolumn{5}{c}{\textbf{DREAM-7B}} \\
RoPE   & 16 & 61.3 & 63.4 & 66.0 \\
MDLMPE & 16 & 61.8 & 63.2 & 67.1 \\
RoPE   & 32 & 61.7 & 62.8 & 65.1 \\
MDLMPE & 32 & 61.9 & 63.3 & \textbf{67.7} \\
RoPE   & 64 & 60.9 & 63.1 & 66.6 \\
MDLMPE & 64 & \textbf{62.0} & \textbf{63.3} & 65.9 \\
\midrule
\multicolumn{5}{c}{\textbf{LLaDA-8B}} \\
RoPE   & 16 & 60.1 & 60.7 & 68.4 \\
MDLMPE & 16 & \textbf{60.4} & 60.1 & 68.4 \\
RoPE   & 32 & 59.5 & 60.2 & 67.4 \\
MDLMPE & 32 & 59.9 & 60.8 & 67.9 \\
RoPE   & 64 & 59.6 & 60.0 & 68.2 \\
MDLMPE & 64 & 60.3 & \textbf{60.9} & \textbf{68.6} \\
\bottomrule
\end{tabular}
\captionof{table}{Matched block-size results. B denotes block size; all scores are accuracies (\%). ARC denotes ARC-Challenge. Within each model panel, data, decoding budget, and evaluation harness are fixed.}
\label{tab:block_sft}
\end{minipage}
\end{figure*}

Beyond the training regime, blockwise generation changes the visibility structure that MDLMPE is designed to represent. Varying the active block determines which response positions are jointly available and thus directly tests topology sensitivity. The two visibility patterns are illustrated in Figure~\ref{fig:block_access}, while each comparison in Table~\ref{tab:block_sft} shares its checkpoint, prompt, block rule, denoising budget, and extractor. Block size changes $r_{ij,f}$ through the mask state; MDLMPE receives no block ID and leaves the update rule unchanged.

Under this controlled change in visibility, the block-size sweep provides concrete evidence across both backbones. For DREAM-7B at block size 32, HellaSwag increases from 65.1 to 67.7 and ARC-Challenge from 62.8 to 63.3. At block size 64, DREAM reaches 62.0 on MMLU and 63.3 on ARC-Challenge. LLaDA-8B at block size 64 improves MMLU from 59.6 to 60.3, ARC-Challenge from 60.0 to 60.9, and HellaSwag from 68.2 to 68.6. Because the same encoder is used without a block identifier, these improvements show that MDLMPE responds directly to the visibility pattern produced by different block configurations.

\subsection{Component Analysis}

Finally, we connect the end-to-end gains back to the structure of the method by examining how its components combine. The incremental configurations in Table~\ref{tab:component} introduce binary availability, Gaussian locality, the shifted-sine basis, and the embedding path in sequence. The reported final row corresponds to the complete dual-path MDLMPE method; the intermediate rows characterize component interactions rather than implying that every added component improves the metric in isolation.

\begin{table}[!t]
\centering
\footnotesize
\setlength{\tabcolsep}{2.2pt}
\begin{tabular}{cccc|cc}
\toprule
\textbf{Bin.} & \textbf{Gauss.} & \textbf{Basis} & \textbf{Emb.} & \textbf{Train loss}$\downarrow$ & \textbf{Eval PPL}$\downarrow$ \\
\midrule
-- & -- & -- & -- & 3.159 & 17.24 \\
\checkmark & -- & -- & -- & 3.193 & 17.67 \\
\checkmark & \checkmark & -- & -- & 3.154 & 16.88 \\
\checkmark & \checkmark & \checkmark & -- & 3.163 & 17.05 \\
\checkmark & \checkmark & \checkmark & \checkmark & \textbf{3.131} & \textbf{16.13} \\
\bottomrule
\end{tabular}
\caption{Component analysis during LLaDA-7B instruction post-training on 50K Tulu examples, evaluated on 2K examples. The first row is RoPE~\cite{su2024roformer}; the final row is the dual-path MDLMPE method.}
\label{tab:component}
\end{table}

The progression in Table~\ref{tab:component} provides a direct link between the representation design and its optimization behavior. Under matched data order, seed, and checkpoint rules, the Gaussian configuration obtains a training loss of 3.154 and an evaluation perplexity of 16.88. The complete dual-path configuration achieves the strongest reported values, with a training loss of 3.131 and an evaluation perplexity of 16.13. These results support combining target-relative Gaussian structure, the multiscale basis, and the two feature-injection paths in the main method. Appendix~E further links the aggregate ARC, MMLU, and GSM8K metrics to paired generations.

\section{Conclusion}

This work introduced MDLMPE, which, to our knowledge, is the first distribution-aware positional encoding method designed for masked diffusion language models. To address the dynamically changing and non-contiguous distributions of revealed and masked positions during denoising, MDLMPE builds upon RoPE to represent token availability through binary availability states, Gaussian local weighting, and a RoPE-aligned sinusoidal basis transformation. The resulting features are incorporated into token embeddings and selected rotary phases. Experiments on LLaDA and DREAM, including supervised fine-tuning of 7B/8B models and pretraining and zero-shot evaluation at the 100M scale, demonstrate that MDLMPE generally outperforms RoPE. These results indicate that dynamic token distributions provide MDLMs with useful positional cues beyond absolute position indices, highlighting the research value and practical potential of MDLMPE. Future work will further evaluate its pretraining effectiveness and scalability on larger MDLMs, explore more effective approaches to token-position-distribution modeling, positional encoding, and feature injection, and examine its generalization across a broader range of model architectures and data distributions.

\let\MDLMPEStandaloneEnd\relax
\ifdefined\MDLMPECombined
\else
\bibliography{references}
\def\MDLMPEStandaloneEnd{\end{document}}
\fi
\MDLMPEStandaloneEnd

\clearpage

\makeatletter
\let\MDLMPEOriginalDocumentClass\documentclass
\let\MDLMPEOriginalUsePackage\usepackage
\let\MDLMPEOriginalTitle\title
\let\MDLMPEOriginalAuthor\author
\let\MDLMPEOriginalAffiliations\affiliations
\let\MDLMPEOriginalMakeTitle\maketitle
\let\MDLMPEOriginalDocument\document
\let\MDLMPEOriginalEndDocument\enddocument
\let\MDLMPEOriginalBibliography\bibliography
\def\documentclass{\@ifnextchar[{\MDLMPEGobbleDocumentClassOpt}{\MDLMPEGobbleDocumentClass}}
\def\MDLMPEGobbleDocumentClassOpt[#1]#2{}
\def\MDLMPEGobbleDocumentClass#1{}
\def\usepackage{\@ifnextchar[{\MDLMPEGobbleUsePackageOpt}{\MDLMPEGobbleUsePackage}}
\def\MDLMPEGobbleUsePackageOpt[#1]#2{}
\def\MDLMPEGobbleUsePackage#1{}
\def\title#1{}
\def\author#1{}
\def\affiliations#1{}
\def\maketitle{}
\def\document{}
\def\enddocument{}
\def\bibliography#1{}
\documentclass[letterpaper]{article} 
\usepackage[preprint]{mdlmpe2027}
\usepackage[hyphens]{url}
\usepackage{graphicx}
\urlstyle{rm}
\def\UrlFont{\rm}
\usepackage{natbib}
\usepackage{caption}
\frenchspacing

\usepackage{booktabs}
\usepackage{multirow}
\usepackage{amsmath,amssymb,amsfonts}
\usepackage{bm}

\setcounter{secnumdepth}{2}

\title{Supplementary Material for MDLMPE:\\
Distribution Aware Positional Encoding for Masked Diffusion Language Models}
\author{
Tong Ling\textsuperscript{\rm 1},
Hang Lei\textsuperscript{\rm 2},
Feng Xiao\textsuperscript{\rm 2},
Changhui Sun\textsuperscript{\rm 3},
Jiahang Xie\textsuperscript{\rm 4},
Hao Liu\textsuperscript{\rm 2},
Lu Liu\textsuperscript{\rm 2},
Yanlong Du\textsuperscript{\rm 2}
}
\affiliations{
\textsuperscript{\rm 1}University Chinese Academic of Science\\
\textsuperscript{\rm 2}HUJING Digital Media \& Entertainment Group\\
\textsuperscript{\rm 3}State Key Laboratory for Novel Software Technology, Nanjing University, China\\
\textsuperscript{\rm 4}School of Data Science, Fudan University, Shanghai, China
}

\begin{document}

\maketitle

\section*{Supplementary Material}

\section*{Overview}

\begingroup
\small
\begin{itemize}
    \item \textbf{Appendix A: Detailed formulation.}
    \begin{itemize}
        \item \textbf{A.1 RoPE-aligned sinusoidal coordinates} formalizes the revealed-token indicator and the paired sinusoidal reference aligned with the RoPE frequency grid.
        \item \textbf{A.2 Embedding-level injection} characterizes residual-stream conditioning by availability-distribution features, its complementarity to rotary attention, and its information-independence and stability properties.
        \item \textbf{A.3 RoPE-aligned sinusoidal basis} examines the shared frequency grid, linear projection, spectral properties, and associated limitations of the fixed coordinate basis.
        \item \textbf{A.4 Local availability ratios and rotary injection} derives the symmetric Gaussian kernel, boundary-normalized mask-conditioned aggregation, pair availability ratio, and bounded pair-conditioned phase adaptation.
        \item \textbf{A.5 Safeguards, variants, and complexity} specifies mixed-precision safeguards, initialization and frequency-allocation strategies, cacheable quantities, and asymptotic computational costs.
    \end{itemize}
    \item \textbf{Appendix B: Extended experimental protocol.}
    \begin{itemize}
        \item \textbf{B.1 Data and availability states} specifies the evaluated MDLM families, parameter scales, training corpora, and construction of denoising-state availability indicators.
        \item \textbf{B.2 Datasets and benchmarks} documents the eight downstream benchmarks and the corpora used for perplexity evaluation.
        \item \textbf{B.3--B.4 Training controls} details the architectures, optimization configurations, hardware settings, and decoding protocols for scratch training and supervised fine-tuning.
        \item \textbf{B.5 Metrics and reporting} defines the positional baselines, accuracy and perplexity metrics, ablation controls, and systems-level measurements.
    \end{itemize}
    \item \textbf{Appendix C: Additional ablations.}
    \begin{itemize}
        \item \textbf{C.1--C.3} examine the Gaussian-window ratio, Gaussian bandwidth, and the respective contributions of the embedding projection and sinusoidal basis.
    \end{itemize}
    \item \textbf{Appendix D: Efficiency analysis.}
    \begin{itemize}
        \item Appendix D derives analytical bounds on the computational and storage overhead of MDLMPE.
    \end{itemize}
    \item \textbf{Appendix E: Qualitative output comparisons.}
    \begin{itemize}
        \item Three tables present paired ARC, MMLU, and GSM8K examples, followed by a consolidated qualitative analysis.
    \end{itemize}
    \item \textbf{Appendix F: Deterministic mechanism analysis.} Formula-aligned deterministic evaluations establish the exact availability-ratio envelope, frequency--distance transfer, local Jacobian bound, intermediate-coordinate sensitivity, and boundary-normalization invariants independently of any sampled token-recovery trajectory or empirical performance claim.
\end{itemize}
\endgroup

\appendix

\section{Detailed Formulation}
\label{app:derivation}

\subsection{RoPE-Aligned Sinusoidal Coordinates}
\label{app:coordinates}

For a batch element with length $L$, let $x_t$ be the token at position $t$, $x_{\mathrm{mask}}$ the mask token, and $a_t$ the attention-validity indicator. The revealed-token indicator is
\begin{equation}
b_t=\mathbb{I}[x_t\neq x_{\mathrm{mask}}]a_t, \qquad b_t\in\{0,1\}.
\label{eq:binary}
\end{equation}
For head dimension $d_h$, there are $F=d_h/2$ rotary pairs. We use RoPE's inverse-frequency grid, $\omega_f=\theta^{-2f/d_h}$, and retain the paired sinusoidal reference coordinate
\begin{equation}
\bm p_{t,f}=\bigl[\sin(t\omega_f),\ \cos(t\omega_f)\bigr].
\label{eq:sinpair}
\end{equation}
The main availability coordinate is the nonnegative, phase-shifted form $c_{t,f}=\frac{1}{2}[1+\sin(t\omega_f)]$. The paired expression documents the underlying RoPE geometry, while the implemented ratio uses the nonnegative coordinate. The next two subsections motivate embedding injection and the shared RoPE spectrum.

\subsection{Embedding-Level Conditioning of Availability Distributions}
\label{app:embedding_injection}

Let $\bm e_i\in\mathbb{R}^{d}$ denote the token embedding at position $i$, $\bm u_i\in\mathbb{R}^{F}$ the unprojected availability aggregate in Eq.~\ref{eq:linear_basis_projection}, and $\bm z_i=\bm W_p\bm u_i+\bm b_p\in\mathbb{R}^{d}$. We inject the projected feature through a gated residual path,
\begin{equation}
\widetilde{\bm e}_i=\bm e_i+\gamma\bm z_i,
\label{eq:embedding_injection}
\end{equation}
where $0\leq\gamma\leq\gamma_{\max}=0.10$ is a learned bounded gate initialized to $0.01$. If $\|\bm z_i\|_2\leq B_z$, then the deviation from the pretrained input satisfies
\begin{equation}
\bigl\|\widetilde{\bm e}_i-\bm e_i\bigr\|_2
=\gamma\|\bm z_i\|_2
\leq\gamma_{\max}B_z.
\label{eq:embedding_perturbation_bound}
\end{equation}
Thus, setting $\gamma=0$ exactly recovers the baseline embedding, while near-zero initialization makes the initial adaptation a controlled perturbation rather than a replacement of the pretrained residual stream.

Embedding injection makes this state information available to every first-layer linear path:
\begin{equation}
\begin{bmatrix}
\bm q_i\\ \bm k_i\\ \bm v_i
\end{bmatrix}
=
\begin{bmatrix}
\bm W_Q\\ \bm W_K\\ \bm W_V
\end{bmatrix}
\widetilde{\bm e}_i
=
\begin{bmatrix}
\bm W_Q\\ \bm W_K\\ \bm W_V
\end{bmatrix}
\bm e_i
+\gamma
\begin{bmatrix}
\bm W_Q\\ \bm W_K\\ \bm W_V
\end{bmatrix}
\bm z_i.
\label{eq:qkv_availability_access}
\end{equation}
Rotary injection directly changes query--key geometry, whereas Eq.~\ref{eq:qkv_availability_access} also exposes the local state to value and subsequent feed-forward paths. Consequently, two denoising states with the same token and absolute index but different revealed neighborhoods can receive different inputs before attention chooses a key.

The residual complements rather than duplicates the \texttt{[MASK]} embedding. Let $y_t$ be the underlying clean token and let $\mathcal{B}(b_{1:L})$ denote the Gaussian aggregation and projection that produces $\bm z_i$. Since it uses only the binary state in Eq.~\ref{eq:binary},
\begin{equation}
\bm z_i=\mathcal{B}(b_{1:L}),\qquad
\mathcal{B}(b_{1:t-1},0,b_{t+1:L})
\ \text{is invariant to }y_t.
\label{eq:availability_content_independence}
\end{equation}
The feature therefore supplies availability geometry, while $\bm e_i$ and the content-dependent projections retain token semantics. In particular, it cannot encode the identity of an unrevealed target; the same construction is applied to the current corruption state during training and to the current denoising state during inference.

The possible adverse effects are correspondingly bounded. A transient pattern changes only the residual term in Eq.~\ref{eq:embedding_injection}; the bounded gate, normalized aggregation, and per-step recomputation let optimization attenuate it. Confusing availability with semantic importance would require the binary feature to replace content, which Eq.~\ref{eq:availability_content_independence} excludes and Eq.~\ref{eq:qkv_availability_access} does not impose. Finally, ordinary RoPE pairs and zero-initialized phase residuals preserve explicit compatibility paths. Appendix~C tests whether this residual information improves results beyond matched parameter counts.

\subsection{RoPE-Aligned Spectral Basis for Availability Projection}
\label{app:sinusoidal_basis}

For frequency $f$, the main coordinate is a nonnegative phase-shifted cosine, equivalently written as a shifted sine,
\begin{equation}
\begin{aligned}
c_{t,f}&=\frac{1}{2}\bigl[1+\cos(t\omega_f-\tfrac\pi2)\bigr]\\
&=\frac{1}{2}\bigl[1+\sin(t\omega_f)\bigr],
\end{aligned}
\qquad 0\leq c_{t,f}\leq1,
\label{eq:nonnegative_coordinate}
\end{equation}
where $\omega_f=\theta^{-2f/d_h}$ is precisely the frequency used by the corresponding RoPE pair. Given the normalized nonnegative weights from Eq.~\ref{eq:normalize}, the availability mass and its all-position reference are
\begin{equation}
\begin{aligned}
A_{i,f}&=\sum_{t=1}^{L}w_{i,t}b_tc_{t,f},\\
D_{i,f}&=\sum_{t=1}^{L}w_{i,t}a_tc_{t,f}+\varepsilon,
\qquad r_{i,f}=\frac{A_{i,f}}{D_{i,f}}.
\end{aligned}
\label{eq:availability_ratio_explicit}
\end{equation}
Because $b_t\in\{0,1\}$ and $w_{i,t},c_{t,f}\geq0$, the numerator is bounded by the unclamped denominator:
\begin{equation}
0\leq A_{i,f}\leq\sum_{t=1}^{L}w_{i,t}a_tc_{t,f}\leq D_{i,f},
\qquad 0\leq r_{i,f}\leq1.
\label{eq:ratio_bound}
\end{equation}
This is why the main ratio uses a shifted coordinate instead of a signed sinusoid: signed contributions could cancel, so a small aggregate would not distinguish scarce revealed evidence from balanced positive and negative phases.

Stacking the coordinates gives $\bm c_t=[c_{t,0},\ldots,c_{t,F-1}]^\top$. The feature before and after the learned linear projection is
\begin{equation}
\bm u_i=\sum_{t=1}^{L}w_{i,t}b_t\bm c_t, \qquad
\bm z_i=\bm W_p\bm u_i+\bm b_p.
\label{eq:linear_basis_projection}
\end{equation}
Hence a one-position change at $t$ has the exact influence
\begin{equation}
\frac{\partial\bm u_i}{\partial b_t}=w_{i,t}\bm c_t,
\qquad
\left\|\frac{\partial\bm u_i}{\partial b_t}\right\|_2
\leq w_{i,t}\sqrt{F}.
\label{eq:single_state_influence}
\end{equation}
The Gaussian weight therefore localizes the effect of a state change, while the fixed vector $\bm c_t$ makes $\bm W_p$ a learned mixture of multiscale spatial summaries. Low-frequency channels describe broad availability imbalance; higher-frequency channels respond to local revealed--masked transitions. The computation remains a weighted vector sum and does not materialize an $L\times L\times F$ tensor.

The alignment with RoPE is structural rather than cosmetic. The original displacement phase $(j-i)\omega_f$, basis coordinate $c_{t,f}$, pair coordinate $s_{ij,f}$, and final residual $\delta_{ij,f}$ are indexed by the same frequency ordering. Thus the model preserves the spectrum represented by a RoPE-pretrained checkpoint instead of introducing an unrelated position scale. It also permits the availability coordinates and Gaussian buffers to reuse the RoPE frequency cache.

Using a periodic basis does not make the encoding the sole source of absolute position. Let $\bm R(\alpha)$ denote the standard $2\times2$ rotation by angle $\alpha$, and write $\alpha_{ij,f}=(j-i)\omega_f$. For any selected rotary pair, the distance between the baseline and residual-conditioned rotation matrices is bounded by
\begin{equation}
\begin{aligned}
\left\|\bm R(\alpha_{ij,f}+\delta_{ij,f})-\bm R(\alpha_{ij,f})\right\|_2
&=2\left|\sin\!\left(\frac{\delta_{ij,f}}{2}\right)\right|\\
&\leq2\sin(\tau_{\max}/2).
\end{aligned}
\label{eq:rotation_perturbation_bound}
\end{equation}
Multiple frequencies, local support, and the retained ordinary RoPE pairs resolve periodic ambiguity, while Eq.~\ref{eq:rotation_perturbation_bound} applies directly to the main pair-conditioned residual with $\tau_{\max}=0.25$ radians.

Alignment can correlate the residual with the original phase, and a fixed basis is less flexible than a fully learned coordinate map. These are controlled by the unchanged RoPE pairs, the bounded residual, and the shared cross-frequency phase MLP. The embedding projection $\bm W_p$ is separate from this phase MLP. Appendix~C compares the fixed and learnable alternatives with matched accounting; the selected basis is therefore a bounded inductive bias with an explicit baseline path rather than an irreversible change to positional geometry.

\subsection{Local Availability Ratios and Rotary Injection}
\label{app:ratios}

Let $W$ be the window radius and $\sigma=\max(\rho W,1)$, where the defaults are $W/L=3/16$ and $\rho=1/4$. Before normalization, the symmetric weight from target position $i$ to source position $t$ is
\begin{equation}
\begin{aligned}
\widetilde w_{i,t}
&=\exp\!\left[-\frac{(i-t)^2}{2\sigma^2}\right]\\
&\quad\times\mathbb{I}[|i-t|\le W]a_t,
\end{aligned}
\label{eq:gaussian}
\end{equation}
where $a_t$ removes padding or otherwise invalid attention positions. The main method has no learned left/right scalar. The row-normalized matrix is
\begin{equation}
w_{i,t}=\frac{\widetilde w_{i,t}}{\sum_{u=1}^{L}\widetilde w_{i,u}+\varepsilon}.
\label{eq:normalize}
\end{equation}
Normalization makes a ratio insensitive to positions retained solely because of boundary truncation. Masked positions are suppressed before the nonnegative basis aggregation in Eq.~\ref{eq:availability_ratio_explicit}. The token-centered matrices $\bm A$ and $\bm D$ are reused by both branches. For an attention pair $(i,j)$, the rotary branch forms the symmetric feature
\begin{equation}
\left[\bm r_{ij}\right]_f
=\frac{A_{i,f}+A_{j,f}}{D_{i,f}+D_{j,f}},
\qquad 0\leq [\bm r_{ij}]_f\leq1.
\label{eq:pair_availability_ratio}
\end{equation}
This ratio is equivalent to averaging the two token-centered numerator sequences and denominator references before division. It is mapped to a centered intermediate coordinate by
\begin{equation}
\bm s_{ij}=\arccos\!\left(2\bm r_{ij}-\bm1\right)-\tfrac\pi2\bm1.
\label{eq:mdangle}
\end{equation}
The intermediate vector is fused across frequencies by the shared MLP,
\begin{equation}
\begin{aligned}
\bm\psi_{ij}&=\operatorname{SiLU}\!\left(\bm W_1\bm s_{ij}+\bm b_1\right),\\
\bm\delta_{ij}&=\tau_{\max}\tanh\!\left(\bm W_2\bm\psi_{ij}+\bm b_2\right),
\end{aligned}
\label{eq:phase_mlp}
\end{equation}
where $\tau_{\max}=0.25$, the hidden width is $H=\max(16,F/2)$, and $(\bm W_2,\bm b_2)$ are initialized to zero. The angular relation used by attention is
\begin{equation}
\Delta\phi_{ij,f}=\begin{cases}
(j-i)\omega_f, & f\in\mathcal F_{\mathrm{RoPE}},\\
(j-i)\omega_f+\delta_{ij,f}, & f\in\mathcal F_{\mathrm{MDLMPE}}.
\end{cases}
\label{eq:app_pair_phase}
\end{equation}
The two frequency sets alternate and each contains approximately half of the rotary pairs. The implementation stores the token-centered $\bm A,\bm D\in\mathbb R^{L\times F}$ summaries once per example; pair ratios and MLP outputs are evaluated in query blocks inside attention, so the full $L\times L\times F$ pair tensor is not materialized.

\subsection{Safeguards, Variants, and Complexity}
\label{app:safeguards}

The ratio denominator is clamped by $\varepsilon$, and the argument of $\arccos$ is clipped away from $\pm1$ before mixed-precision evaluation. These safeguards prevent division by zero and endpoint-gradient instabilities. The zero-initialized final MLP layer makes $\bm\delta_{ij}=\bm0$ at initialization, exactly recovering RoPE in the rotary branch. A learnable nonnegative affine basis can replace the fixed coordinate for an ablation, but the main method keeps the aligned fixed basis. With a full window, aggregation costs $O(L^2F)$; a radius-$W$ grouped-convolution or banded reduction costs $O(LWF)$. The enabled embedding projection costs $O(LFd)$, and pair-conditioned phase evaluation costs $O(L^2FH)$. Geometry buffers depend only on $(L,W,\rho,\theta)$ and are cached; the token-centered state is computed once per model forward and reused by all layers, while pair phases are evaluated in query blocks with $O(CLF)$ extra working storage for query-block size $C$.

\section{Extended Experimental Protocol}
\label{app:protocol}

\subsection{Data and Availability-State Construction}

We evaluate LLaDA~\cite{nie2025llada} and DREAM~\cite{ye2025dream} at two scales: matched 100M models trained from scratch and RoPE-pretrained LLaDA-8B/DREAM-7B checkpoints adapted by SFT. Pretraining uses OpenWebText (OWT)~\cite{gokaslan2019openwebtext}; SFT uses Tulu-3~\cite{lambert2024tulu3} and SmolTalk~\cite{allal2025smollm2smolgoesbig}. All ablations use LLaDA-7B on Tulu SFT with 50{,}000 training examples and 2{,}000 test examples. All training experiments use $8\times$ NVIDIA A100 80GB GPUs. Each example is tokenized once by the native tokenizer of the evaluated checkpoint, then padded, truncated, and corrupted under the same rule for all compared encodings. At a denoising step, the availability indicator is reconstructed from the actual model input: prompt tokens and revealed response tokens have value one, masked response tokens have value zero, and padding positions are excluded by the attention-validity mask. This avoids treating a nominal diffusion timestep as a substitute for the realized revealed/masked arrangement.

\subsection{Datasets and Evaluation Benchmarks}

Main-paper Table~1 groups eight downstream benchmarks into three categories. General tasks comprise MMLU~\cite{hendrycks2021measuring}, MMLU-Pro~\cite{wang2024mmlupro}, HellaSwag~\cite{zellers2019hellaswag}, and ARC-Challenge~\cite{clark2018arc}. Math and reasoning tasks comprise GSM8K~\cite{cobbe2021training} and Hendrycks MATH~\cite{hendrycks2021math}. Code tasks comprise HumanEval~\cite{chen2021codex} and MBPP~\cite{austin2021program}.

Main-paper Table~2 reports perplexity on the One Billion Word Benchmark (1BW)~\cite{chelba2014one}, Penn Treebank (PTB)~\cite{marcus1993building}, Text8~\cite{mahoney2011text8}, and WikiText-2 and WikiText-103~\cite{merity2017pointer}, with OWT~\cite{gokaslan2019openwebtext} used for pretraining and in-domain validation. These references identify the data sources; all compared encodings use identical preprocessing and evaluation within each matched experiment.

\subsection{Matched Scratch-Training Protocol}

The 100M models use hidden size 768, 12 layers, 12 attention heads, four key-value heads, a 126{,}464-token vocabulary, RMSNorm, SiLU, tied embeddings, and maximum length 1{,}024. OWT pretraining runs for 200K steps with learning rate $1\times10^{-4}$, effective batch size 16, AdamW ($\beta_1=0.9$, $\beta_2=0.999$, weight decay 0.01), cosine decay, and 2K learning-rate-warmup steps. This optimizer schedule does not activate the MDLMPE head-level warmup used for SFT; MDLMPE is active from the first pretraining update. The RoPE baseline and every MDLMPE variant start from the same architecture and initialization protocol. A matched run changes only the position module; data order, token budget, optimizer states, sequence length, corruption sampler, and validation cadence remain fixed. This control is important because a change in mask sampling can itself change diffusion loss even when the model weights are unchanged.

\subsection{SFT Adaptation and Decoding Controls}

SFT uses learning rate $2\times10^{-5}$, five epochs, prompt-loss masking, group-by-length batches, and bfloat16 FSDP. RoPE and MDLMPE start from the same source checkpoint and use the same backbone training policy, learning-rate scheduler, stopping rule, and checkpoint-selection criterion. The projection, gate, and phase-MLP parameters are present only in MDLMPE and are included in its parameter accounting. As part of checkpoint-compatible MDLMPE integration, its rotary heads are activated progressively; this method-specific head schedule does not alter the optimizer, data order, token budget, or decoding controls. For total update count $N$ and update $n$, the active MDLMPE-head fraction is
\begin{equation}
\alpha(n)=
\begin{cases}
0, & n/N<0.1,\\
\dfrac{n/N-0.1}{0.8}, & 0.1\le n/N<0.9,\\
1, & n/N\ge0.9.
\end{cases}
\label{eq:sft_head_warmup}
\end{equation}
Thus SFT begins with RoPE, linearly activates MDLMPE on key--value heads between 10\% and 90\% of training, and finishes with MDLMPE active on all heads. For an active count $h$, the selected key--value indices are approximately uniform over $[0,H_{\mathrm{KV}}-1]$ rather than always taking a prefix; grouped query heads inherit the state of their associated key--value head. The warmup is disabled for 100M scratch pretraining, where MDLMPE is active throughout, and evaluation/inference uses all heads. Prompt template, candidate order, generation budget, denoising schedule, seed policy, and answer normalization are fixed at evaluation. The availability feature is recomputed after each iterative update from the current corrupted sequence.

\subsection{Metrics and Evaluation Harness}

RoPE~\cite{su2024roformer} is the primary baseline; ALiBi~\cite{press2022alibi} and absolute embeddings are diagnostics at 100M. Perplexity is computed with each model's native diffusion objective under matched masking schedules, reporting the same token subset for every encoding. Multiple-choice accuracy uses an identical prompt template, candidate serialization, decoding budget, answer parser, and invalid-output treatment within a matched comparison. The benchmark suite covers language modeling and knowledge, scientific, arithmetic, and commonsense decisions; PPL is therefore interpreted together with task accuracy rather than as a sufficient proxy for downstream utility.

\subsection{Reporting, Ablations, and Systems Measurement}

Component and hyperparameter ablations retain the same parameter accounting unless the purpose is explicitly to test added learnable capacity. Throughput and memory are measured only after warmup with fixed sequence length, batch size, dtype, attention implementation, device, and denoising steps; peak allocated memory is reset before each measurement, and throughput is averaged over a fixed interval. A configuration is described as robust only when its direction agrees across the matched language-modeling and task-level metrics, while a quality--systems trade-off is reported as such rather than being folded into a single score.

\section{Additional Ablations}
\label{app:ablations}

\begingroup
\setlength{\intextsep}{2pt plus 1pt minus 1pt}
\setlength{\textfloatsep}{2pt plus 1pt minus 1pt}
\setlength{\floatsep}{2pt plus 1pt minus 1pt}
\setlength{\abovecaptionskip}{3pt}
\setlength{\belowcaptionskip}{0pt}

All studies use LLaDA-7B on the Tulu SFT split with 50{,}000 training and 2{,}000 test examples and report training loss and evaluation perplexity.

\subsection{Window Ratio}

The window ratio governs global coverage versus local emphasis. Holding other parameters fixed, we compare progressively local $W/L$ settings in Table~\ref{tab:app_window}.

\begin{table}[!htbp]
\centering
\small
\setlength{\tabcolsep}{4pt}
\begin{tabular}{c|cc}
\toprule
\textbf{Window ratio} & \textbf{Train loss}$\downarrow$ & \textbf{Eval PPL}$\downarrow$ \\
\midrule
1 & 3.152 & 16.77 \\
3/4 & 3.155 & 16.72 \\
1/2 & 3.143 & 16.59 \\
1/4 & 3.136 & 16.38 \\
3/16 & \textbf{3.131} & \textbf{16.13} \\
1/8 & 3.134 & 16.26 \\
\bottomrule
\end{tabular}
\caption{Gaussian-window-ratio ablation.}
\label{tab:app_window}
\end{table}
\FloatBarrier

Table~\ref{tab:app_window} indicates that a window ratio of $3/16$ yields the lowest training loss and evaluation perplexity among the tested settings, while more global and more local windows both incur higher perplexity.

\subsection{Gaussian Standard Deviation}

The Gaussian bandwidth governs positional decay within the retained window. We vary $\rho$ in $\sigma=\rho W$ while holding $W$ fixed; Table~\ref{tab:app_sigma} compares concentrated and smoother local weighting.

\begin{table}[!htbp]
\centering
\small
\setlength{\tabcolsep}{4pt}
\begin{tabular}{c|cc}
\toprule
$\bm\rho$ & \textbf{Train loss}$\downarrow$ & \textbf{Eval PPL}$\downarrow$ \\
\midrule
0.20 & 3.142 & 16.33 \\
0.25 & \textbf{3.131} & \textbf{16.13} \\
0.33 & 3.132 & 16.21 \\
0.50 & 3.144 & 16.40 \\
\bottomrule
\end{tabular}
\caption{Gaussian-bandwidth ablation.}
\label{tab:app_sigma}
\end{table}
\FloatBarrier

Table~\ref{tab:app_sigma} identifies $\rho=0.25$ as the best setting among the tested bandwidths for both reported metrics.

\subsection{Embedding Projection and Sinusoidal Basis}

The RoPE-aligned sinusoidal basis and embedding projection $\bm W_p$ may contribute independently. We therefore evaluate a $2\times2$ factorial design while retaining the same binary state, Gaussian weights, and pair-conditioned rotary branch. When the sinusoidal basis is disabled, the positional coordinate is represented by an MLP rather than omitted. Table~\ref{tab:app_linear_sin} reports the four configurations.

\begin{table}[!htbp]
\centering
\small
\setlength{\tabcolsep}{3pt}
\begin{tabular}{cc|cc}
\toprule
\textbf{Emb. proj.} & \textbf{Basis} & \textbf{Train loss}$\downarrow$ & \textbf{Eval PPL}$\downarrow$ \\
\midrule
No & No & 3.160 & 17.18 \\
Yes & No & 3.134 & 16.42 \\
No & Yes & 3.163 & 17.05 \\
Yes & Yes & \textbf{3.131} & \textbf{16.13} \\
\bottomrule
\end{tabular}
\caption{Embedding-projection and sinusoidal-basis ablation.}
\label{tab:app_linear_sin}
\end{table}
\FloatBarrier

Table~\ref{tab:app_linear_sin} shows that the configuration combining the embedding projection and sinusoidal basis attains the lowest training loss and evaluation perplexity among the evaluated settings.

\endgroup
\section{Efficiency Analysis}
\label{app:efficiency}

This section derives the computational and storage complexity of the pair-conditioned implementation. The analysis separates cached geometric quantities from denoising-state-dependent tensors and avoids reliance on empirical systems measurements.

\subsection{Cached Local Aggregation}

Let $\bm C\in\mathbb{R}^{L\times F}$ contain the nonnegative basis coordinates and let $\bm K_W\in\mathbb{R}^{L\times L}$ denote the conceptual sparse, row-normalized truncated Gaussian operator. At one denoising state, the token-centered numerator and reference matrices are
\begin{equation}
\begin{aligned}
\bm A&=\bm K_W\operatorname{diag}(\bm b)\bm C,\\
A_{i,f}&=\sum_{|t-i|\leq W}K_{i,t}b_tc_{t,f},\\
D_{i,f}&=\sum_{|t-i|\leq W}K_{i,t}a_tc_{t,f}+\varepsilon.
\end{aligned}
\label{eq:efficiency_local_aggregation}
\end{equation}
The support of $\bm K_W$ gives $\operatorname{nnz}(\bm K_W)\leq L(2W+1)$ and hence
\begin{equation}
\begin{aligned}
T_{\mathrm{local}}&=O(LWF),\\
T_{\mathrm{embed}}&=O(LFd),\\
T_{\mathrm{pair}}&=O(L^2FH),\\
T_{\mathrm{add}}&=T_{\mathrm{local}}+T_{\mathrm{embed}}+T_{\mathrm{pair}}.
\end{aligned}
\label{eq:efficiency_time_bound}
\end{equation}
The first term is the local reduction, the second is the enabled embedding projection, and the third evaluates $\bm r_{ij}$ and the shared cross-frequency phase MLP for all attention pairs. The Gaussian kernel, basis, and frequency assignment depend only on $(L,W,\rho,\theta)$ and are cached. Padding-aware $\bm A$ and $\bm D$ vary with the denoising state and are computed once per forward pass. A grouped one-dimensional convolution or banded reduction evaluates the local operator, while query-block evaluation of $\bm r_{ij}$ avoids materializing the full $L\times L\times F$ pair tensor.

\subsection{Runtime Decomposition}

For $N_\ell$ unchanged transformer layers, write the per-step costs as
\begin{equation}
\begin{aligned}
T_{\mathrm{RoPE}}&=N_\ell T_{\mathrm{block}}+T_{\mathrm{pos}},\\
T_{\mathrm{MDLMPE}}&=N_\ell T_{\mathrm{block}}+T_{\mathrm{pos}}
+T_{\mathrm{local}}+T_{\mathrm{embed}}+T_{\mathrm{pair}}.
\end{aligned}
\label{eq:efficiency_runtime_decomposition}
\end{equation}
MDLMPE does not change layer count, hidden size, attention-mask shape, denoising steps, or backbone query/key/value dimensions. Because the residual phase is pair conditioned, however, the attention kernel requires an MDLMPE-aware extension; the reference implementation evaluates it in query blocks. The relative per-step overhead is
\begin{equation}
\frac{T_{\mathrm{MDLMPE}}-T_{\mathrm{RoPE}}}{T_{\mathrm{RoPE}}}
=\frac{T_{\mathrm{local}}+T_{\mathrm{embed}}+T_{\mathrm{pair}}}
{N_\ell T_{\mathrm{block}}+T_{\mathrm{pos}}}.
\label{eq:efficiency_relative_runtime}
\end{equation}
This expression identifies the additional per-step computation attributable to MDLMPE.

\subsection{Activation and Buffer Bounds}

The query, key, value, output, and KV-cache backbone shapes are identical to RoPE, while pair-conditioned phase evaluation adds block-local state inside attention. With query-block size $C$, the additional storage obeys
\begin{equation}
\begin{aligned}
M_{\mathrm{persistent}}&=O(W+LF+Ld),\\
M_{\mathrm{pair\mbox{-}block}}&=O(CLF),
\end{aligned}
\label{eq:efficiency_memory_bound}
\end{equation}
where the $O(Ld)$ projected feature can be streamed or reused after embedding injection. The binary vector $\bm b$ is input derived, and the fixed kernel and basis need no gradient storage. Thus
\begin{equation}
M_{\mathrm{MDLMPE}}=M_{\mathrm{RoPE}}+O(W+LF+Ld+CLF).
\label{eq:efficiency_total_memory}
\end{equation}
An unblocked implementation has $C=L$ and therefore requires $O(L^2F)$ pair working storage. Query blocking controls this term but does not remove the pair-phase computation. Any systems evaluation should specify $L$, $W$, $C$, dtype, selected-frequency count, attention kernel, batch size, and denoising steps.

\FloatBarrier

\section{Qualitative Output Comparisons}
\label{app:demos}

Tables~\ref{tab:demo_arc}--\ref{tab:demo_gsm8k} present nine paired cases from ARC, MMLU, and GSM8K. Across these examples, the MDLMPE responses provide complete and internally consistent solutions for the selected instances. The comparisons contrast non-terminating or self-contradictory baseline outputs in ARC, invalid algebraic premises in MMLU, and omitted quantities or arithmetic errors in GSM8K with responses whose stated reasoning supports the final answer. These examples offer qualitative evidence concerning response completion and reasoning consistency; they do not constitute an aggregate benchmark evaluation.

\begin{table*}[!t]
\centering
\small
\setlength{\tabcolsep}{3pt}
\renewcommand{\arraystretch}{1.03}
\begin{tabular}{p{0.315\textwidth}p{0.285\textwidth}p{0.285\textwidth}}
\toprule
\textbf{Question and reference} & \textbf{RoPE response} & \textbf{MDLMPE response} \\
\midrule
\textbf{Example 1, ARC-148.} How many valence electrons does selenium have? (A) 3; (B) 5; (C) 6; (D) 8. \textbf{Ref.: C.}
& \textit{Selenium is a nonmetal with atomic number 34 and is in group 66. 66. 66. 66. \ldots} The response enters an unbounded numeric repetition, does not recover the intended group number, and never emits the requested final-answer sentence.
& \textit{Selenium has electron configuration $[\mathrm{Ar}]\,3d^{10}4s^24p^4$. The outer shell contains two $4s$ electrons and four $4p$ electrons, for six valence electrons. Equivalently, selenium is a group-16 element. The best answer is C.} \\
\midrule
\textbf{Example 2, ARC-511.} Which object is orbited by a belt of asteroids? (A) Pluto; (B) Saturn; (C) the Sun; (D) the Moon. \textbf{Ref.: C.}
& \textit{The best answer is B, Saturn. The asteroid belt is between Mars and Jupiter and is not directly orbiting Saturn, so the best answer is B.} The continuation repeats the prompt and this internally inconsistent answer several times: its explanation rules out Saturn while its final letter remains B.
& \textit{The main asteroid belt occupies a region between the orbits of Mars and Jupiter, but its members follow heliocentric orbits. It is therefore a belt of objects orbiting the Sun, not a structure orbiting Saturn, Pluto, or the Moon. The best answer is C.} \\
\midrule
\textbf{Example 3, ARC-716.} Which statement describes a physical characteristic of the Moon? (A) made of hot gases; (B) covered with many craters; (C) many bodies of liquid water; (D) gives off its own light. \textbf{Ref.: B.}
& \textit{The Moon is a rocky satellite with many impact craters. It lacks a substantial atmosphere and liquid surface water, so weathering and erosion do not rapidly erase those craters. It is neither made of hot gas nor a source of its own visible light. Therefore, the correct answer is B.}
& \textit{The Moon's surface preserves numerous impact craters formed over billions of years. Its extremely thin exosphere, absence of flowing surface water, and limited geological resurfacing allow these structures to remain visible. The Moon reflects sunlight rather than producing its own light. The best answer is B.} \\
\bottomrule
\end{tabular}
\caption{ARC examples 1--3 with their questions, references, and paired responses.}
\label{tab:demo_arc}
\end{table*}

\begin{table*}[!t]
\centering
\small
\setlength{\tabcolsep}{3pt}
\renewcommand{\arraystretch}{1.03}
\begin{tabular}{p{0.315\textwidth}p{0.285\textwidth}p{0.285\textwidth}}
\toprule
\textbf{Question and reference} & \textbf{RoPE response} & \textbf{MDLMPE response} \\
\midrule
\textbf{Example 1, MMLU-46.} S1: A homomorphism may have an empty kernel. S2: No nontrivial homomorphism can map some finite group into some infinite group. (A) T,T; (B) F,F; (C) T,F; (D) F,T. \textbf{Ref.: B.}
& \textit{Statement 1 is true because an injective homomorphism can have an empty kernel. Statement 2 is false because a finite group can have a nontrivial homomorphism into an infinite group. The best answer is C.} The first step confuses a trivial kernel with an empty kernel.
& \textit{For every homomorphism $\phi:G\to H$, the identity satisfies $\phi(e_G)=e_H$, hence $e_G\in\ker\phi$ and the kernel cannot be empty; S1 is false. S2 is also false because an infinite group may contain a finite subgroup, allowing a nontrivial inclusion, for example $C_2\hookrightarrow C_2\times\mathbb Z$. The best answer is B.} \\
\midrule
\textbf{Example 2, MMLU-61.} S1: Integer $2\times2$ matrices with nonzero determinant form a group. S2: Integer $2\times2$ matrices with determinant one form a group. (A) T,T; (B) F,F; (C) T,F; (D) F,T. \textbf{Ref.: D.}
& \textit{Both sets satisfy closure, associativity, identity, and inverse axioms, because every matrix with nonzero determinant is invertible. Therefore both statements are true and the best answer is A.} This overlooks that invertibility over $\mathbb Q$ does not guarantee an inverse with integer entries.
& \textit{S1 is false: $\operatorname{diag}(2,1)$ has nonzero determinant but inverse $\operatorname{diag}(1/2,1)$, which is not an integer matrix. S2 is true because a determinant-one integer matrix has inverse equal to its integer adjugate. Thus the determinant-one matrices form $\mathrm{SL}(2,\mathbb Z)$. The best answer is D.} \\
\midrule
\textbf{Example 3, MMLU-91.} Find all cosets of $4\mathbb Z$ in $2\mathbb Z$: (A) $4\mathbb Z$; (B) $4\mathbb Z,\,2+4\mathbb Z$; (C) $2\mathbb Z$; (D) $\mathbb Z$. \textbf{Ref.: B.}
& \textit{Elements of $2\mathbb Z$ not in $4\mathbb Z$ are represented by 2 or $-2$. Adding either representative to $4\mathbb Z$ gives the same second coset. Thus the cosets are $4\mathbb Z$ and $2+4\mathbb Z$, so the best answer is B.}
& \textit{Every even integer is congruent either to 0 or to 2 modulo 4. The first class is $4\mathbb Z$; the second is $2+4\mathbb Z$, and $-2+4\mathbb Z$ is identical to that second class. Hence $[2\mathbb Z:4\mathbb Z]=2$ and there are exactly two cosets. The best answer is B.} \\
\bottomrule
\end{tabular}
\caption{MMLU examples 1--3 with the decisive algebraic condition in each response.}
\label{tab:demo_mmlu}
\end{table*}

\begin{table*}[!t]
\centering
\small
\setlength{\tabcolsep}{3pt}
\renewcommand{\arraystretch}{1.03}
\begin{tabular}{p{0.315\textwidth}p{0.285\textwidth}p{0.285\textwidth}}
\toprule
\textbf{Question and reference} & \textbf{RoPE response} & \textbf{MDLMPE response} \\
\midrule
\textbf{Example 1, GSM8K-798.} When Billy was first hired, he earned \$10 per hour. After two months he received a \$0.50 raise, and on his first anniversary he received a \$1.00 raise. Sally's starting wage is \$0.50 above Billy's original starting wage. If both work 20 hours, how much more does Billy earn? \textbf{Ref.: \$20.}
& \textit{To solve this problem, calculate Billy's and Sally's hourly wages and then determine their total earnings for 20 hours. Let's break it down step by step. First, calculate Billy's \ldots} The output then restarts the complete problem, begins the same first step again, restarts a second time, and truncates. It never combines the two raises, computes either 20-hour total, or emits the required final-answer line.
& \textit{Billy's current hourly wage is $\$10.00+\$0.50+\$1.00=\$11.50$. Sally earns $\$10.00+\$0.50=\$10.50$ per hour because her comparison is with Billy's original wage. For 20 hours, Billy earns $20\times\$11.50=\$230$ and Sally earns $20\times\$10.50=\$210$. The difference is $\$230-\$210=\$20$. The final answer is \$20.} \\
\midrule
\textbf{Example 2, GSM8K-997.} Walter collects \$500 from his neighbors and \$1,500 from an online fund. His lawyer offers to donate three times as much as everyone else donates. Before Walter tells the lawyer the total, a friend gives another \$200. How much will the lawyer contribute? \textbf{Ref.: \$6,600.}
& \textit{Walter first collects $\$500+\$1{,}500=\$2{,}000$. His lawyer offers three times everyone else's donation, so $3\times\$2{,}000=\$6{,}000$. Therefore Walter's lawyer is going to contribute \$6000. The final answer is \$6000.} This calculation is internally consistent for the first two sources but drops the friend's \$200 even though it arrives before Walter reports the total.
& \textit{The multiplier applies to the donations made by everyone other than the lawyer. Those sources contribute $\$500$ from neighbors, $\$1{,}500$ online, and $\$200$ from the friend, for $\$500+\$1{,}500+\$200=\$2{,}200$. Three times that complete total is $3\times\$2{,}200=\$6{,}600$. The final answer is \$6600.} \\
\midrule
\textbf{Example 3, GSM8K-1014.} Jamaar goes to the store with \$10 earned from mowing lawns. Including tax, peaches and pears cost \$0.50 each, apples cost \$0.75, kiwis cost \$1, and plums cost \$0.25. After buying 3 peaches, 4 pears, 2 kiwis, and 5 apples, how many plums can he buy? \textbf{Ref.: 3.}
& \textit{The cost of peaches is listed as \$1, pears as \$3, kiwis as \$1, and apples as \$3. These amounts total \$8, leaving $\$10-\$8=\$2$. At \$0.25 per plum, $\$2/\$0.25=8$, so Jamaar can buy 8 plums. The final answer is 8.} The subtraction and division are valid given the subtotal, but four preceding multiplication terms are copied or computed incorrectly.
& \textit{Compute each category independently: $3\times\$0.50=\$1.50$ for peaches, $4\times\$0.50=\$2.00$ for pears, $2\times\$1.00=\$2.00$ for kiwis, and $5\times\$0.75=\$3.75$ for apples. The subtotal is $\$1.50+\$2.00+\$2.00+\$3.75=\$9.25$, leaving $\$10-\$9.25=\$0.75$. Therefore $\$0.75/\$0.25=3$ plums. The final answer is 3.} \\
\bottomrule
\end{tabular}
\caption{GSM8K examples 1--3 with their questions, references, and paired responses.}
\label{tab:demo_gsm8k}
\end{table*}

\section{Deterministic Mechanism Analysis}
\label{app:visualizations}

The figures in this section are direct evaluations of Eqs.~\ref{eq:availability_ratio_explicit}, \ref{eq:single_state_influence}, \ref{eq:normalize}, and~\ref{eq:mdangle}. They use analytical extrema or explicitly fixed positions; no recovery schedule, random token placement, checkpoint output, attention map, or benchmark result enters any panel. Figures~\ref{fig:app_exact_envelope} and~\ref{fig:app_frequency_distance} use the default $W/L=3/16$ and $\sigma/W=1/4$ at length 1024. The length-256 Jacobian and boundary panels use $W=64$ and $\sigma=16$ only to make the full matrices legible; their locality and normalization claims do not depend on choosing the default radius.

\subsection{Exact Availability Envelope}

For a fixed target and frequency, define the normalized contribution of source position $t$ as
\begin{equation}
\alpha_{i,t,f}=\frac{w_{i,t}c_{t,f}}{\sum_u w_{i,u}c_{u,f}},
\qquad
r_{i,f}=\sum_t\alpha_{i,t,f}b_t,
\label{eq:normalized_source_contribution}
\end{equation}
where $\alpha_{i,t,f}\geq0$ and $\sum_t\alpha_{i,t,f}=1$ in the idealized $\varepsilon\to0$ limit. Let $N_i$ be the number of valid positions in the local support and let $\eta=k/N_i$ denote the revealed fraction within that support. If exactly $k$ positions are revealed, the largest attainable ratio is obtained by selecting the $k$ largest $\alpha_{i,t,f}$ values, and the smallest by selecting the $k$ smallest. These sorted partial sums give exact extrema over every binary mask with that cardinality; no enumeration or stochastic recovery path is required.

\begin{figure}[!htbp]
\centering
\includegraphics[width=0.98\columnwidth]{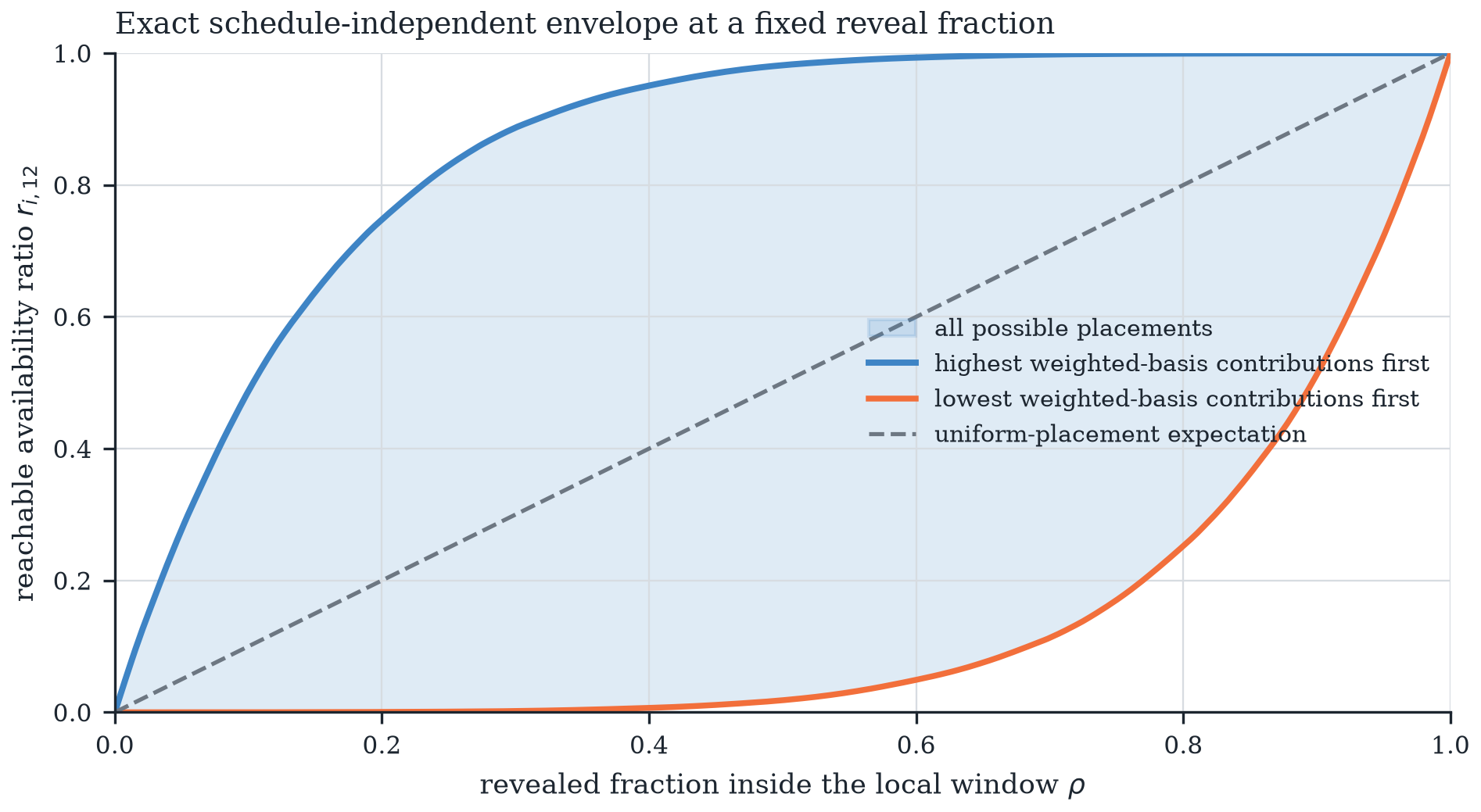}
\caption{Exact schedule-independent availability envelope at rotary pair $f=12$, with fixed radius $W=192$ and $\sigma=48$. At each local revealed fraction $\eta$, the shaded band contains every attainable $r_{i,12}$. Its upper and lower boundaries reveal the positions with the largest and smallest weighted-basis contributions first; the diagonal is the uniform-placement expectation, not a sampled trajectory.}
\label{fig:app_exact_envelope}
\end{figure}

Figure~\ref{fig:app_exact_envelope} separates density from topology. Fixing $\eta$ does not fix $r_{i,f}$: the vertical width of the envelope is precisely the range created by changing revealed-token placement while preserving the number of revealed positions. The envelope collapses to zero at $\eta=0$ and to one at $\eta=1$, while its nonzero interior width proves that intermediate availability ratios retain spatial information beyond global density.

\subsection{Frequency--Distance Transfer}

\begin{figure}[!htbp]
\centering
\includegraphics[width=0.98\columnwidth]{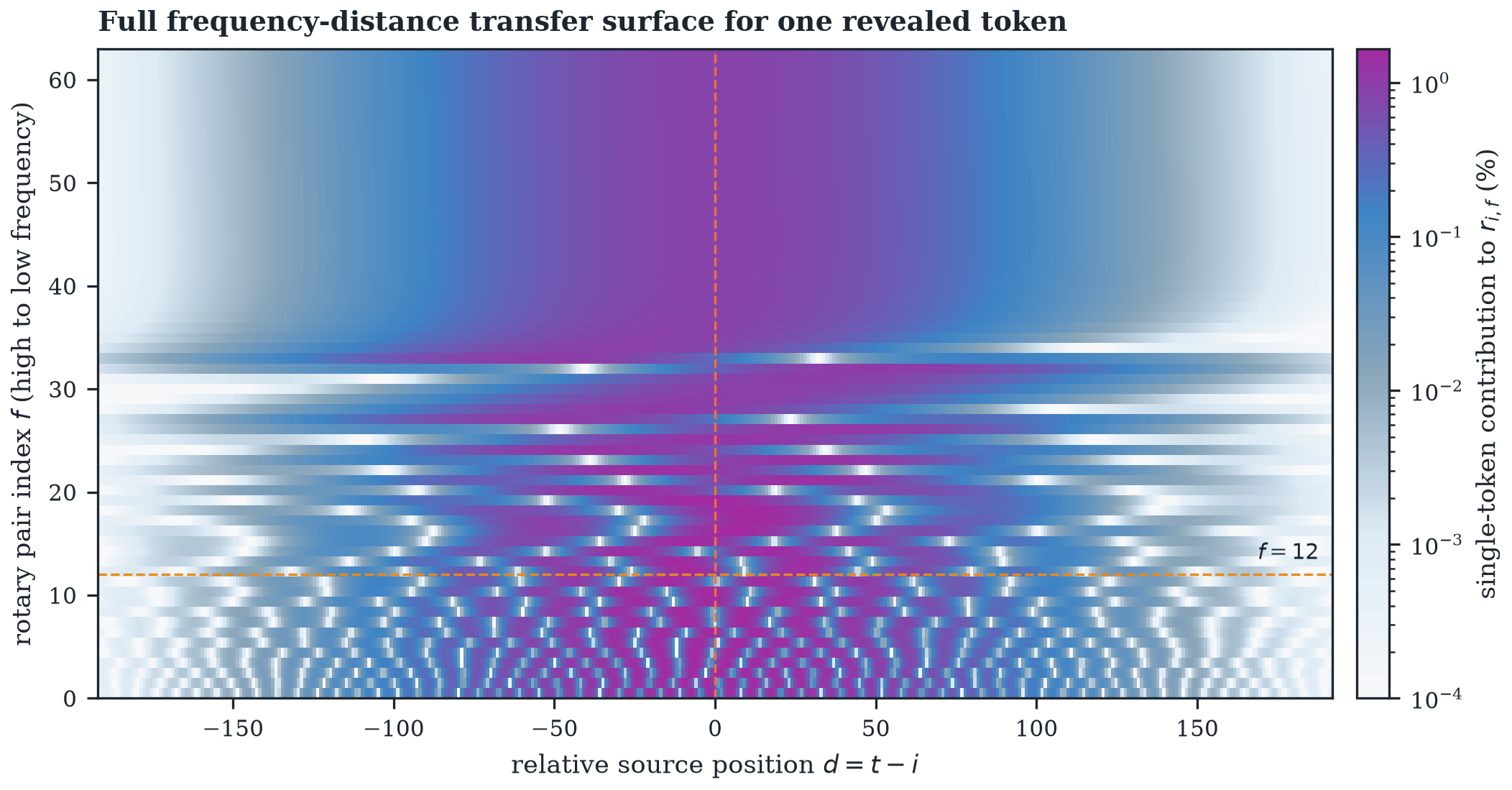}
\caption{Frequency--distance transfer for one revealed source token in a fixed length-1024 sequence, with center target, $W=192$, and $\sigma=48$. Each cell is $100\alpha_{i,t,f}$, the percentage contribution to $r_{i,f}$. The vertical dashed line marks the target and the horizontal dashed line marks $f=12$. Gaussian locality creates the bounded horizontal support; the RoPE-aligned basis produces the frequency-dependent internal structure.}
\label{fig:app_frequency_distance}
\end{figure}

Starting from an all-masked state and revealing only source $t$ changes the ratio at target $i$ by $\alpha_{i,t,f}$. Evaluating this quantity for every relative offset and every rotary pair yields Figure~\ref{fig:app_frequency_distance}. The Gaussian factor confines influence to $|t-i|\leq W$, whereas the aligned coordinate $c_{t,f}$ redistributes that local mass across frequencies. The result is a complete deterministic transfer surface rather than a selection of illustrative token positions.

The surface clarifies the division of labor between the kernel and the spectral basis. Distance controls whether a source can contribute and its broad attenuation, while frequency controls how the same source is represented across channels. Thus the fixed construction supplies multiple spatial scales without assigning a learned semantic score to any position.

\subsection{Local Influence and Its Analytical Bound}

\begin{figure}[!htbp]
\centering
\includegraphics[width=0.98\columnwidth]{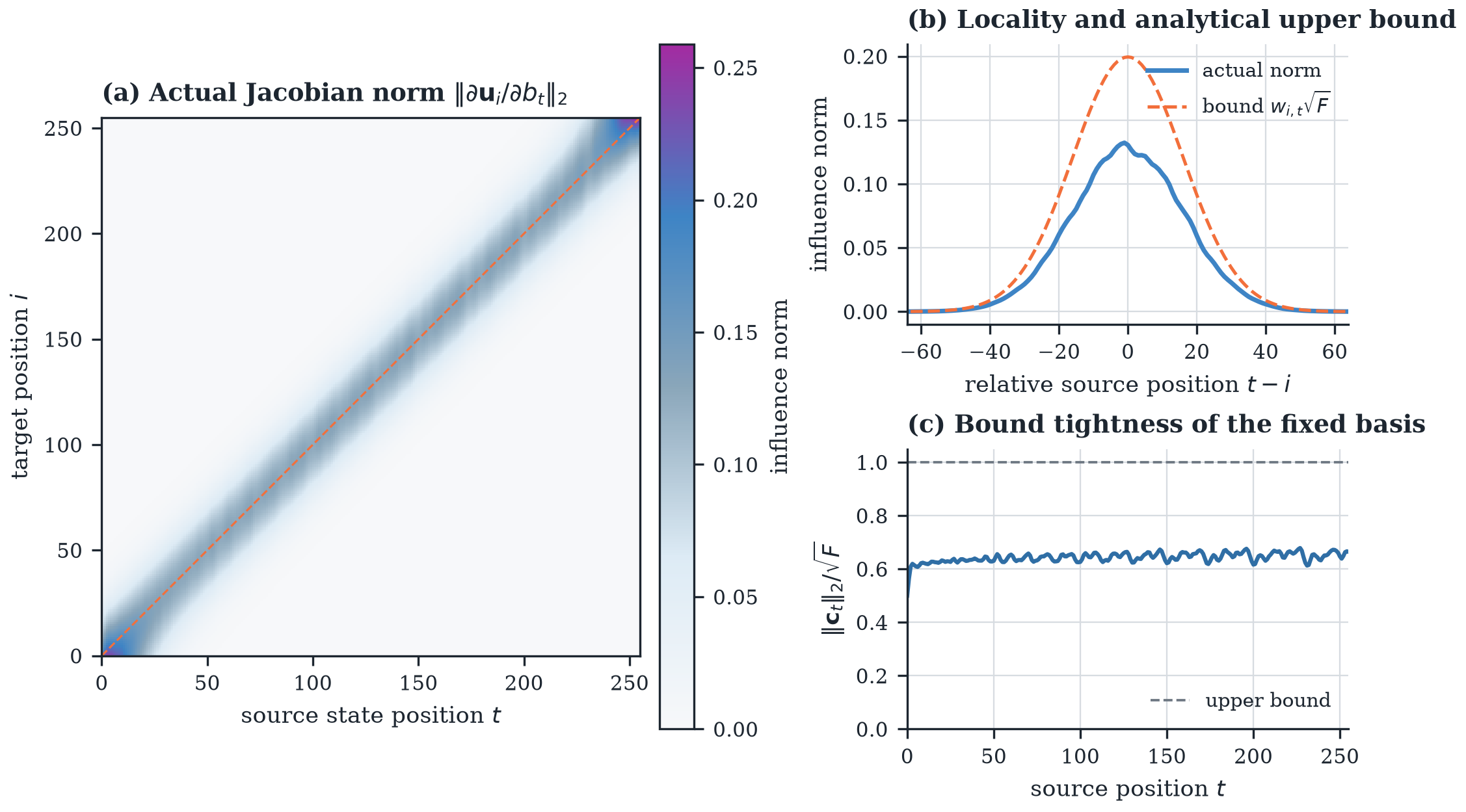}
\caption{Exact single-state influence and analytical localization bound for length 256, $F=64$, $W=64$, and $\sigma=16$. (a) The full Jacobian-norm matrix is banded around $t=i$. (b) A center-target cross-section compares the exact norm with $w_{i,t}\sqrt{F}$. (c) The basis factor $\|\bm c_t\|_2/\sqrt{F}$ never exceeds one, verifying the bound independently of a token-recovery pattern.}
\label{fig:app_jacobian_bound}
\end{figure}

Equation~\ref{eq:single_state_influence} gives the exact Jacobian of the unprojected multiscale feature with respect to one binary state. Because each component of $\bm c_t$ lies in $[0,1]$, $\|\bm c_t\|_2\leq\sqrt{F}$ and therefore
\begin{equation}
\left\|\frac{\partial\bm u_i}{\partial b_t}\right\|_2
=w_{i,t}\|\bm c_t\|_2
\leq w_{i,t}\sqrt{F}.
\label{eq:jacobian_local_bound}
\end{equation}
Figure~\ref{fig:app_jacobian_bound} evaluates both sides for all target--source pairs under one fixed configuration.

The banded matrix establishes strict locality: outside the truncated Gaussian support, changing $b_t$ has zero direct effect on $\bm u_i$. Inside the support, the dashed bound isolates the worst-case magnitude contributed by the fixed basis. The plot therefore makes both the receptive field and the maximum per-token perturbation auditable before learning $\bm W_p$.

\subsection{Intermediate Angular Coordinate and Endpoint Sensitivity}

\begin{figure}[!htbp]
\centering
\includegraphics[width=0.98\columnwidth]{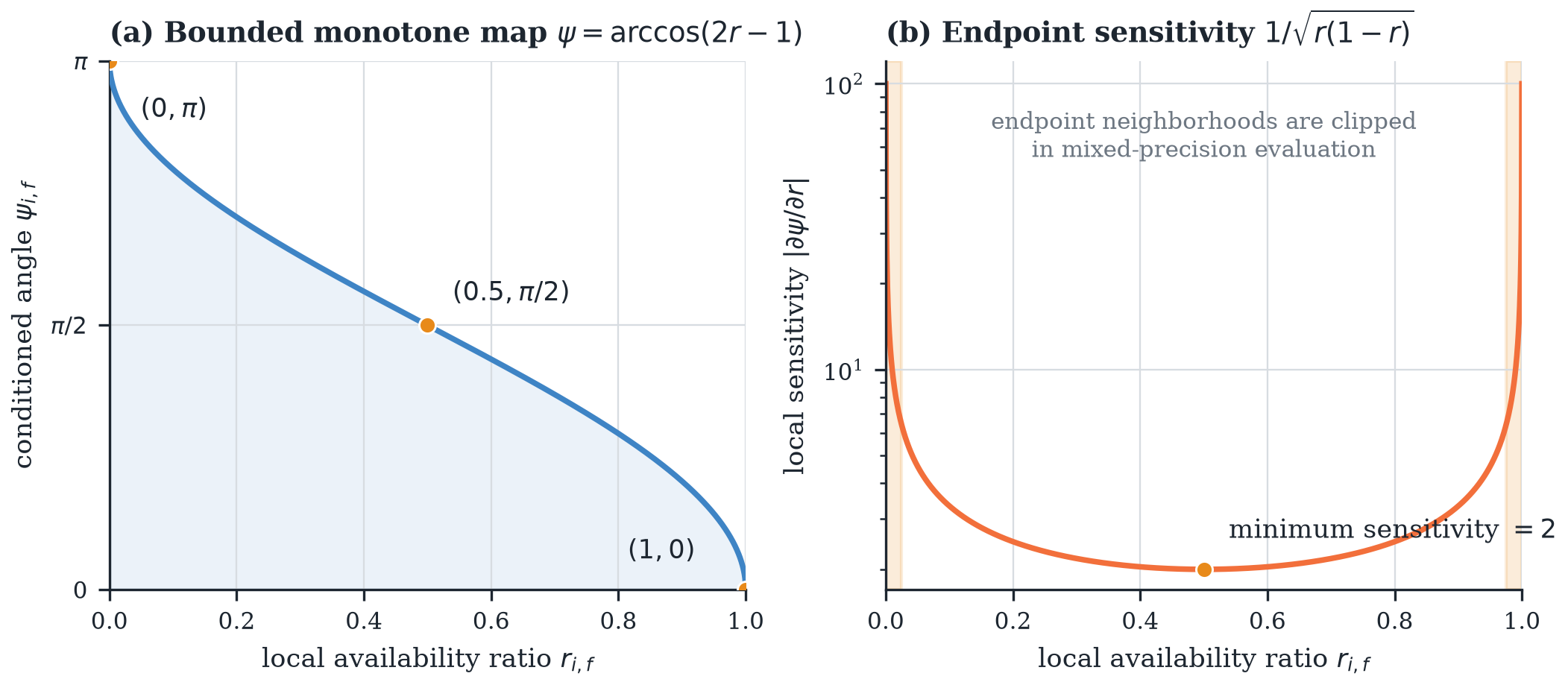}
\caption{Deterministic ratio-to-intermediate-coordinate conversion. (a) The uncentered map $q=\arccos(2r-1)$ maps $r=0,1/2,1$ to $\pi,\pi/2,0$; the implemented pair coordinate is $s=q-\pi/2$, which is passed to the shared MLP and is not substituted directly for RoPE. (b) The exact local sensitivity $|\partial q/\partial r|=|\partial s/\partial r|=1/\sqrt{r(1-r)}$ is smallest at $r=1/2$ and grows near both endpoints; the shaded endpoint neighborhoods indicate where mixed-precision clipping is applied.}
\label{fig:app_ratio_angle}
\end{figure}

For scalar analysis, let $q(r)=\arccos(2r-1)$ denote the uncentered angularization and $s(r)=q(r)-\pi/2$ the centered coordinate used in Eq.~\ref{eq:mdangle}. The first map is monotone from $[0,1]$ to $[0,\pi]$, while centering shifts its range to $[-\pi/2,\pi/2]$ without changing its derivative:
\begin{equation}
\frac{\partial q}{\partial r}
=\frac{\partial s}{\partial r}
=-\frac{1}{\sqrt{r(1-r)}}.
\label{eq:ratio_angle_derivative}
\end{equation}
The magnitude is minimized at $r=1/2$, where it equals two, and diverges as $r$ approaches either endpoint. Figure~\ref{fig:app_ratio_angle} shows the deterministic angularization and its complete sensitivity curve, rather than selected availability states.

The two panels distinguish output range from numerical sensitivity. Bounded $s$ prevents unbounded inputs to the phase MLP, while the subsequent $\tau_{\max}\tanh(\cdot)$ independently bounds the actual pair-conditioned rotary residual. Bounded values alone do not bound the derivative of the intermediate map. Equivalently, clipping the ratio to $\widetilde r=\operatorname{clip}(r,\delta,1-\delta)$ gives the explicit finite bound
\begin{equation}
\left|\frac{\partial s}{\partial\widetilde r}\right|
\leq\frac{1}{\sqrt{\delta(1-\delta)}}.
\label{eq:clipped_angle_bound}
\end{equation}
Thus endpoint clipping controls mixed-precision sensitivity while leaving the interior map unchanged. It is a numerical safeguard near all-masked and all-revealed states, not the final phase rule or a mechanism for selecting a recovery schedule.

\subsection{Boundary Normalization and Endpoint Invariants}

\begin{figure}[!htbp]
\centering
\includegraphics[width=0.98\columnwidth]{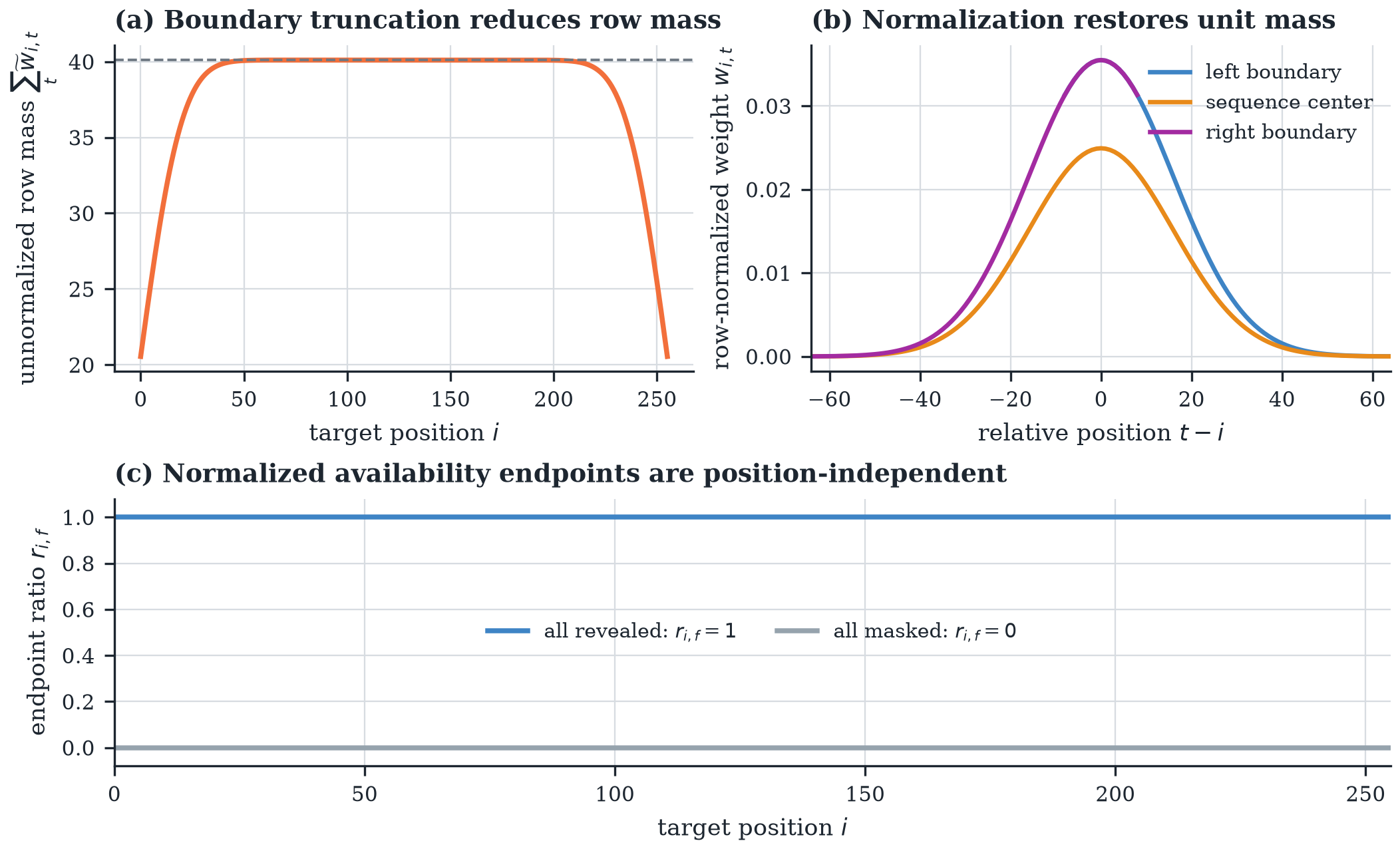}
\caption{Boundary normalization for length 256, $W=64$, and $\sigma=16$. (a) Truncation reduces unnormalized Gaussian row mass near sequence ends. (b) Renormalized kernels at the left boundary, center, and right boundary each have unit mass despite asymmetric support. (c) Consequently, the ideal endpoint ratios are position independent: $r_{i,f}=0$ for an all-masked state and $r_{i,f}=1$ for an all-revealed state (up to the denominator safeguard $\varepsilon$).}
\label{fig:app_boundary_invariants}
\end{figure}

Sequence boundaries truncate a symmetric Gaussian window, so the unnormalized row mass is smaller near the first and last positions. Row normalization in Eq.~\ref{eq:normalize} removes this purely geometric loss of support. Figure~\ref{fig:app_boundary_invariants} compares the raw mass, normalized kernels, and the two uniform endpoint states.

This invariance prevents a boundary position from appearing less available merely because half of its nominal neighborhood lies outside the sequence. Boundary effects can still change the distribution of relative offsets in panel (b), but they do not alter the meaning of the two availability endpoints. Together, Figures~\ref{fig:app_exact_envelope}--\ref{fig:app_boundary_invariants} establish range, topology sensitivity, multiscale locality, bounded influence, angular sensitivity, and edge behavior. They characterize the representation; task-quality claims continue to rely on the matched experiments in the main paper and Appendices~B--D.

\FloatBarrier

\bibliography{references}

\let\documentclass\MDLMPEOriginalDocumentClass
\let\usepackage\MDLMPEOriginalUsePackage
\let\title\MDLMPEOriginalTitle
\let\author\MDLMPEOriginalAuthor
\let\affiliations\MDLMPEOriginalAffiliations
\let\maketitle\MDLMPEOriginalMakeTitle
\let\document\MDLMPEOriginalDocument
\let\enddocument\MDLMPEOriginalEndDocument
\let\bibliography\MDLMPEOriginalBibliography
\makeatother

{\fontsize{8}{9}\selectfont
\bibliography{references}
}
\end{document}